\documentclass{bmvc2k}

\usepackage{url}
\usepackage{graphicx}

\usepackage{amsmath}
\usepackage{amssymb}
\usepackage{amsfonts}
\usepackage{newtxtext,newtxmath}

\usepackage{subcaption}
\usepackage{adjustbox}
\usepackage{color, colortbl}

\usepackage{booktabs}       
\usepackage{nicefrac}       
\usepackage{microtype}     
\usepackage{xcolor}        
\usepackage{multirow}
\usepackage{multicol}
\usepackage{xspace}
\usepackage{adjustbox}
\usepackage{import}
\usepackage{subcaption}
\usepackage{color, colortbl}
\usepackage{makecell}
\usepackage{wrapfig}
\usepackage{enumitem}
\usepackage{pifont}
\usepackage{threeparttable}

\usepackage{color}
\usepackage{xcolor}
\definecolor{citecolor}{HTML}{0071bc}
\definecolor{mlpMixerColor}{RGB}{242,122,130}
\definecolor{concatColor}{RGB}{0,118,186}
\definecolor{dwconvColor}{RGB}{254,174,0}

\newcommand{\xmark}{\ding{55}}

\title{PlainMamba: Improving Non-Hierarchical Mamba in Visual Recognition}

\addauthor{\small Chenhongyi Yang$^*$}{chenhongyi.yang@ed.ac.uk}{1}
\addauthor{\small Zehui Chen$^*$}{lovesnow@mail.ustc.edu.cn}{2}
\addauthor{\small Miguel Espinosa$^*$}{miguel.espinosa@ed.ac.uk}{1}
\addauthor{\small Linus Ericsson}{linus.ericsson@ed.ac.uk}{1}
\addauthor{\small Zhenyu Wang}{2301210449@stu.pku.edu.cn}{3}
\addauthor{\small Jiaming Liu}{jiamingliu@stu.pku.edu.cn}{3}
\addauthor{\small Elliot J. Crowley}{elliot.j.crowley@ed.ac.uk}{1}

\addinstitution{
 \small School of Engineering,\\
 University of Edinburgh
}
\addinstitution{
 \small University of Science and Technology of China
}
\addinstitution{
 \small Peking University
}
\runninghead{C. Yang}{PlainMamba}

\begin{document}

\maketitle
\newcommand{\ourmethod}{PlainMamba}

\newcommand{\apbox}{AP$^{bb}$}
\newcommand{\apmask}{AP$^{mk}$}

\begin{abstract}
We present \ourmethod: a simple non-hierarchical state space model (SSM) designed for general visual recognition. The recent Mamba model has shown how SSMs can be highly competitive with other architectures on sequential data and initial attempts have been made to apply it to images. In this paper, we further adapt the selective scanning process of Mamba to the visual domain, enhancing its ability to learn features from two-dimensional images by (i) a \textit{continuous 2D scanning} process that improves spatial continuity by ensuring adjacency of tokens in the scanning sequence, and (ii) \textit{direction-aware updating} which enables the model to discern the spatial relations of tokens by encoding directional information.
Our architecture is designed to be easy to use and easy to scale, formed by stacking identical \ourmethod~blocks, resulting in a model with constant width throughout all layers. The architecture is further simplified by removing the need for special tokens. We evaluate \ourmethod~on a variety of visual recognition tasks, achieving performance gains over previous non-hierarchical models and is competitive with hierarchical alternatives. For tasks requiring high-resolution inputs, in particular, \ourmethod~requires much less computing while maintaining high performance. Code and models are
available at: \url{https://github.com/ChenhongyiYang/PlainMamba}.

\end{abstract}

\section{Introduction}
\label{sec:intro}
\vspace{-2mm}

Developing high-performing visual encoders has always been one of the most important goals in computer vision~\cite{resnet,resnext,vgg,ViT_dosovitskiy2021an,DeiT_touvron2020,Swin_Liu_2021tq,cswin}. With high-quality visual features, a broad range of downstream tasks, such as semantic segmentation~\cite{ADK20K, cheng2021mask2former,xiao2018unified, wang2023cdac}, object recognition~\cite{DeiT_touvron2020,resnet,resnext,Liu2022Jan_convnext} and detection~\cite{RetinaNet_Lin_2017_ICCV,MaskRCNN_He_2017_ICCV,NIPS2015_14bfa6bb} can be tackled with relative ease. Early methods for extracting visual representations relied on hand-crafted features such as SIFT ~\cite{lowe2004distinctive_sift} and SURF~\cite{Bay2006_surf}. A big breakthrough then came with the adoption of convolutional neural networks (CNNs) that process images with local contexts and enforce spatial equivariance \cite{alexnet,resnet,vgg}. Recently, vision transformers~(ViTs)~\cite{ViT_dosovitskiy2021an} obviated the need for such enforced inductive biases in favour of learnable contexts that operate on image patches \cite{Transformer_NIPS2017_Vaswani,DeiT_touvron2020,Swin_Liu_2021tq}. However, despite the overwhelming success of transformers and their self-attention mechanism \cite{bert_devlin-etal-2019-bert,gpt3,yang2023gpvit}, the quadratic cost of attention has proved to be an obstacle to further scaling such models.

This has invigorated interest in state space models (SSMs) \cite{lti,mamba,hippo,liu2024vmamba,zhu2024vision}. Due to their close ties to linear recurrent networks, SSMs have the benefits of potentially infinite context lengths while maintaining linear complexity in the input sequence length \cite{mamba}, offering substantial speedups compared to attention. However, it took several notable advances to make SSMs effective at learning competitive representations, including enforcing state space variables to be orthogonal basis projections \cite{hippo}. The recent Mamba~\cite{mamba} architecture further aligned SSM-based models with modern transformers, such as making the state space variables input-dependent --- much like queries, keys and values in self-attention. When being used for NLP, those designs led to a state space model that could scale to the sizes and performances of modern transformer-based LLMs~\cite{gpt3,touvron2023llama}, while improving inference effiency.

\begin{figure*}[!t]
    \centering
    \includegraphics[width=\textwidth]{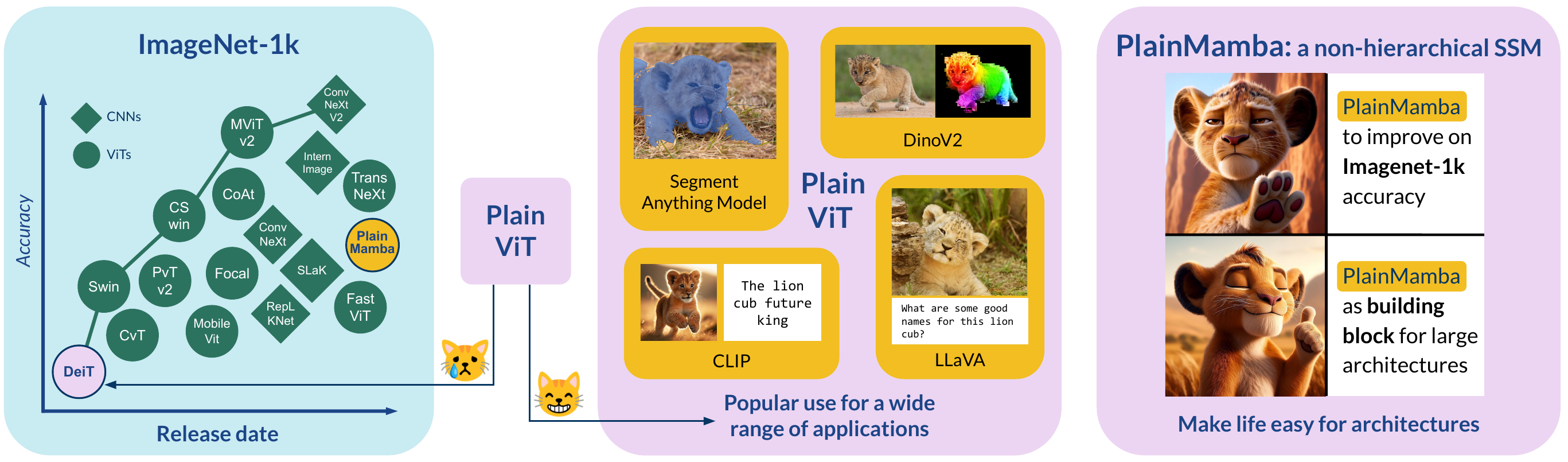}
\vspace{-0.1in}
    \captionsetup{font={footnotesize}}
\caption{While hierarchical visual encoders may demonstrate superior accuracy on open-source visual recognition benchmarks, the plain non-hierarchical models have had more widespread use because of their simple structure. We investigate the potential of the plain Mamba model in visual recognition.}
\label{fig:teaser}
\vspace{-5mm}
\end{figure*}

There is now understandable interest in adapting the Mamba architecture to the visual domain \cite{liu2024vmamba,li2024mamba_nd,zhu2024vision}. However, before we start doing that, we need to think about under what guidelines should we design our new model. As we show in Figure~\ref{fig:teaser}, by examining the development of recently proposed visual encoders, we find that adding more inductive biases, e.g., hierarchical structure, to the plain model such as DeiT can indeed improve a model's performance on open-source benchmarks like ImageNet. However, we should not ignore the fact that the plain ViT~\cite{ViT_dosovitskiy2021an} is widely used by several popular vision foundation models~\cite{sam_model,llava,clip,dinov2,dalle2}, which suggests that simplicity in architecture design is key for multiple reasons. Firstly, maintaining a constant model width (i.e. non-hierarchical) makes it much easier to integrate features from multiple levels, as is common in dense prediction tasks such as semantic segmentation \cite{sam_model}. It also becomes easier to combine features across different modalities such as in CLIP \cite{clip} or LLaVa \cite{llava} or as parts of increasingly complex AI-powered systems. Furthermore, simpler components can be more easily optimized for hardware acceleration~\cite{dao2022flashattention}. In addition, it has also been observed that the over-crafted models may lead to a significant gap between the performance on commonly used benchmarks and downstream tasks \cite{ericsson2021well,donewithimagenet}.
This means benchmark performance may no longer reflect real-world usefulness, as over-engineering tends to increase model complexity and thus make it harder for others to re-use.

Motivated by the above findings, we propose \textbf{\ourmethod}: a simple Mamba architecture for visual recognition. This model integrates ideas from CNNs, Transformers and novel SSM-based models with an aim to providing easy-to-use models for the vision modality. Compared to previous visual state space models \cite{liu2024vmamba,zhu2024vision}, we simplify the architecture by maintaining constant model width across all layers of the network via stacking identical blocks as well as removing the need for \texttt{CLS} tokens. This allows for easy scaling and model re-use, while achieving competitive performances.

Our contributions are as follows: \textbf{(1)} We propose a new visual state space model we call \textit{\ourmethod}. This architecture improves and simplifies previous attempts at extending the Mamba architecture to the visual modality. \textbf{(3)} We improve the SSM block by adapting selective scanning to better process 2D spatial inputs, in two ways. (i) Our \textbf{continuous 2D scanning} approach ensures that the scanning sequence is spatially continuous to improve semantic continuity. (ii) Our \textbf{direction-aware updating}, inspired by positional encoding, allows the model to encode the directionality of each scanning order to further improve spatial context. \textbf{(3)}We test our \textit{\ourmethod} architecture using three different sizes (7M, 26M and 50M) and show how they perform competitively on a range of tasks, from ImageNet1K classification to semantic segmentation and object detection. Specifically, we show that \ourmethod~outperforms its non-heretical counterparts, including SSMs and Transformers, while performing on par with the hierarchical competitors. 

\vspace{-0.1in}
\section{Related Work}
\vspace{-2mm}

\noindent\textbf{Visual Feature Extractors.}
How to effectively extract visual features from images has been a long-standing challenge in computer vision. %
In the early years of deep learning, CNNs~\cite{vgg,alexnet,resnet,resnext} dominated the model architecture landscape. Their induced spatial prior, through the use of convolutional filters, exploits the locality of visual features. Furthermore, stacking multiple layers increases their receptive field. Many different CNN backbone architectures have been proposed over the years~\cite{alexnet,densenet,Chollet2016Oct}, introducing new ways of exploiting spatial information~\cite{vgg, resnext}, building deeper models~\cite{resnet,inception}, improving efficiency~\cite{mobilenetv2,efficientnet_pmlr_tan_19,radosavovic2020designing}, adding multi-scale connections~\cite{unet}, scaling architectures~\cite{convnextv2}, and introducing attention mechanisms~\cite{residual_attention,lambdanetworks_bello2021,BoT_srinivas2021,ECA_wang2020,SENet_Hu_2018,chen2019drop}. In recent years, ViTs have become a powerful tool for image modeling \cite{ViT_dosovitskiy2021an}. Compared to CNNs, they make fewer assumptions about data (feature locality~\cite{xu2021vitae}, translation and scale invariance). By replacing the convolutional layers with self-attention modules, transformers can capture global relationships and have achieved state-of-the-art results on many common image benchmarks~\cite{imagenet_deng2009,COCO_Lin_2014vm,ADK20K}. To adapt the original transformer architecture~\cite{Transformer_NIPS2017_Vaswani} for vision tasks, images are split into patches and converted into tokens before being fed into the transformer encoder. Within this framework, numerous works have focused on pushing the performance (e.g. LeViT,~\cite{LeViT_BenGraham_2021vh} combining transformer encoder layers and convolutions), or on reducing the costly quadratic complexity of self-attention~\cite{flash_attention,flash_attention2}. Another popular extension to ViT architectures has been the addition of hierarchical structures~\cite{PVT_wang2021,Swin_Liu_2021tq,xu2021co,CvT_Wu_2021tw,fan2021multiscale}, similar to the multi-scale feature pyramids used in CNNs. The Swin Transformer~\cite{Swin_Liu_2021tq}, for instance, uses shifted windows to share feature information across scales. These multi-scale features are then used for a wide range of downstream tasks. Recent research has explored ways of using these hierarchical features within ViTs themselves~\cite{stunned,CMT,mvitv2, cswin, gcvit, chen2022regionvit, ConViTdAscoli_2021vz, MPViT,Shi2023Nov_transnext}. Some works~\cite{Swin_Liu_2021tq} have examined the use of multi-resolution features as attention keys and values to learn multi-scale information. However, these extensions add complexity to the model and make it harder to effectively use its features in later stages, thus hindering widespread adoption. Indeed, recent works~\cite{li2022exploring, yang2023gpvit} return to the original ViT architecture, as its non-hierarchical nature greatly simplifies the use of its features. In particular, the plain ViT provides greater flexibility for pre-training and fine-tuning on different tasks. \\

\begin{figure*}[!t]
    \centering
    \includegraphics[width=\textwidth]{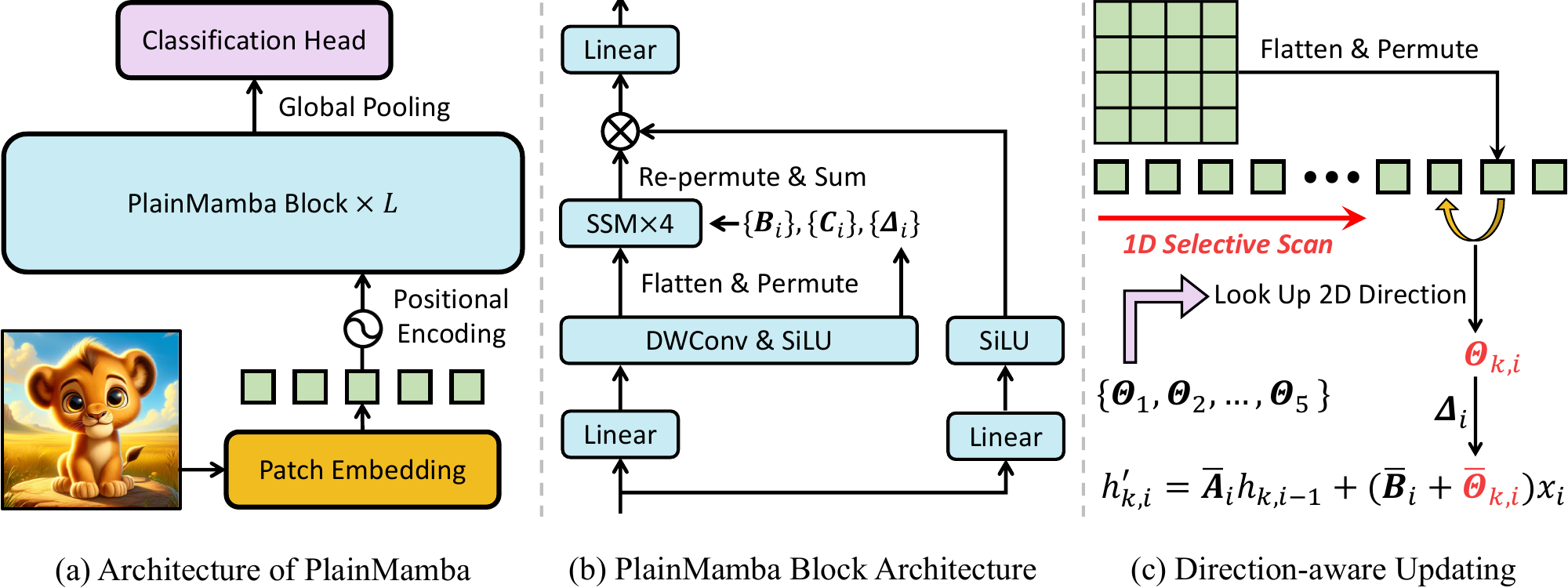}
\vspace{-0.05in}
    \captionsetup{font={footnotesize}}
\caption{(a) The overall architecture of the proposed \ourmethod. \ourmethod~does not have a hierarchical structure, it instead stacks $L$ identical \ourmethod~block to form the main network. For image classification, it uses global average pooling instead of the \texttt{CLS} to gather global information.
(b) Architecture of \ourmethod~block, which is similar to the Mamba~\cite{mamba} block where the selective scanning is combined with a gated MLP. (c) The proposed \textit{Direction-Aware Updating}, where a series of learnable parameters $\mathbf{\Theta}_k$ are combined with the data-dependent updating parameters to explicitly inject relative 2D positional information into the selective scanning process.}
\label{fig:pipeline}
\vspace{-5mm}
\end{figure*}

\noindent\textbf{State Space Models.} 
State Space Models (SSMs) have emerged as efficient alternatives to transformers and CNNs due to their ability to scale linearly with sequence length~\cite{Gu2021NovLSSL,Sun2023Jul}. SSMs transform the state space to effectively capture dependencies over extended sequences. To alleviate the initial computational cost of such models, S4~\cite{Gu2021OctS4} enforced low-rank constraints on the state matrix and S5~\cite{Smith2022AugS5} introduced parallel scanning to further improve efficiency. Furthermore, H3~\cite{Fu2022DecHippo} achieved competitive results on common benchmarks by improving the hardware utilization. Lastly, Mamba~\cite{mamba} parameterized the SSM matrices as functions of the input, thus allowing it to act as a learnable selection mechanism and providing greater flexibility. Follow-up works have extended selective SSMs for images \cite{Baron2023Jun_2d_ssm,Ruan2024Feb_vm_unet,Liu2024Feb_swin_umamba,Wang2024Feb_mamba_unet,Wang2024Feb_semi_mamba_unet,ma2024u,pei2024efficientvmamba,liu2024swin} and videos~\cite{nguyen2022s4nd} using a hierarchical structure~\cite{liu2024vmamba} and bidirectional blocks~\cite{zhu2024vision}, while Mamba-ND~\cite{li2024mamba_nd} introduces an architecture for multi-dimensional data. MambaIR~\cite{Guo2024Feb_mambair} tackles image restoration, and Pan-Mamba~\cite{He2024Feb_panmamba} works on pan-sharpening. DiS~\cite{Fei2024Feb_mamba_diffusion} introduces SSMs to diffusion models by replacing the U-Net with an SSM backbone. While drawing inspiration from the above works, \ourmethod~improves Mamba's~\cite{mamba} selective SSM block by adding wider depth-wise convolutions. In contrast to the Cross-Scan Module (CSM)~\cite{liu2024vmamba} and Mamba-ND~\cite{li2024mamba_nd}, \ourmethod~respects the spatio-sequential nature of image patches (see Figure \ref{fig:pipeline}). As opposed to~\cite{zhu2024vision}, we do not use the \texttt{CLS} token.
\\

\noindent\textbf{Simplifying Visual Feature Extractors.}
Simplifying and unifying existing methods is equally important as improving performance. Plain architectures are robust, conceptually simpler, and scale better. ViTs \cite{ViT_dosovitskiy2021an} remove the pyramid structure of CNNs by converting images into patched tokens. This way, they easily adapt the transformer architecture for visual tasks. Another trick that stems form sequence modeling is the usage of \texttt{CLS} tokens for prediction, which have proven to be unnecessary for visual tasks~\cite{Zhai2021Jun_scaling_vits}. FlexiVit~\cite{Beyer2022Dec_flexivit} unified into a single architecture images with different input resolutions, and GPViT \cite{yang2023gpvit} improved feature resolution with a non-hierarchical transformer. Similarly, ConvNext~\cite{Liu2022Jan_convnext} introduced a simple CNN model that competed with state-of-the-art transformer methods. Other works, like MLP-Mixer~\cite{tolstikhin2021mlp} and follow-up works~\cite{vision_permutator}, have introduced simple architectures using only multi-layer perceptrons. The plain non-hierarchical ViT~\cite{ViT_dosovitskiy2021an} has served as a simple building block for many diverse tasks. SAM~\cite{sam_model} uses a pre-trained ViT as image encoder with minimal changes for image segmentation at large scale. DinoV2~\cite{dinov2,vits_need_registers} uses a ViT to learn general-purpose visual features by pretraining models on curated datasets with self-supervision. Similarly, the image encoder for the CLIP~\cite{clip} model consists of a basic ViT with minor modifications, allowing image-text representations to be learned with a contrastive objective. DALLE-2~\cite{dalle2} incorporates a ViT image encoder to extract visual features that are used for text-conditional image generation.
LlaVA~\cite{llava,improvedllava} combines a vision encoder (pretrained ViT from CLIP) and an LLM for vision-language tasks.

\vspace{-4mm}
\section{Method}
\vspace{-2mm}

\subsection{Preliminaries}
\label{subsec:preliminaries}

\noindent\textbf{State Space Models.} SSMs are typically used to model a continuous linear time-invariant (LTI) system~\cite{lti} where an input signal $x(t) \in \mathbb{R}$ is mapped to its output signal $y(t) \in \mathbb{R}$ through a state variable $h(t) \in \mathbb{R}^{m}$ with the following rules:
\begin{align}
    h'(t) = \mathbf{A}h(t) + \mathbf{B}x(t)\mathrm{,}~~~y(t) = \mathbf{C}h'(t) + \mathbf{D}x(t) 
\end{align}
where $\mathbf{A} \in \mathbb{R}^{m \times m}$, $\mathbf{B} \in \mathbb{R}^{m \times 1}$, $\mathbf{C} \in \mathbb{R}^{1 \times m}$ and  $\mathbf{D} \in \mathbb{R}^{1 \times 1}$ are parameters. To make the above system usable for a discrete system, e.g., a sequence-to-sequence task, a timescale parameter $\mathbf{\Delta}$ is used to transform the parameters $\mathbf{A}$ and $\mathbf{B}$ to their discretized counterparts $\bar{\mathbf{A}}$ and $\bar{\mathbf{B}}$. In Mamba~\cite{mamba} and its following works~\cite{liu2024vmamba,zhu2024vision}, this is achieved with the following zero-order hold (ZOH) rule:
\begin{align}
    \bar{\mathbf{A}} = \exp{(\mathbf{\Delta}\mathbf{A})}\mathrm{,}~~~\bar{\mathbf{B}} = (\mathbf{\Delta}\mathbf{A})^{-1}(\exp{(\mathbf{\Delta}\mathbf{A})}-\mathbf{I}) \cdot \mathbf{\Delta}\mathbf{B}
\end{align}
Afterwards, an input sequence $\{x_i\}$ (for $i= 1, 2, ...$) can be mapped to its output sequence $\{y_i\}$ in a similar way:
\begin{align}
\label{eq:ssm-output}
    h'_i = \bar{\mathbf{A}}h_{i-1} + \bar{\mathbf{B}}x_i\mathrm{,}~~~y_i = \mathbf{C}h'_i + \mathbf{D}x_i 
\end{align}

\noindent\textbf{Mamba.}
Since SSMs are often used to model LTI systems, their model parameters are shared by all time steps $i$. However, as found in Mamba~\cite{mamba}, such time-invariant characteristics severely limit the model's representativity. To alleviate this problem, Mamba lifts the time-invariant constraint and makes the parameters $\mathbf{B}$, $\mathbf{C}$ and $\mathbf{\Delta}$ dependent on the input sequence $\{x_i\}$, a process they refer to as the \textit{selective scan}, resulting in the token-dependent $\{\mathbf{B}_i\}$, $\{\mathbf{C}_i\}$ and $\{\mathbf{\Delta}_i\}$. Moreover, the SSM is combined with a gated MLP~\cite{gated} to gain better representation ability. Specifically, the output sequence $\{y_i\}$ is computed from the $\{x_i\}$ as the following:
\begin{align}
    x'_i = \sigma\bigl(&\mathrm{DWConv}\bigl(\mathrm{Linear}(x_i)\bigr)\bigr)\mathrm{,}~~~z_i = \sigma\bigl(\mathrm{Linear}(x_i)\bigr)\\
    \mathbf{B}_i,\mathbf{C}_i,&\mathbf{\Delta}_i = \mathrm{Linear}(x'_i)\mathrm{,}~~~\bar{\mathbf{A}}_i,\bar{\mathbf{B}}_i=\mathrm{ZOH}(\mathbf{A}, \mathbf{B_i}, \mathbf{\Delta_i}) \\
  h'_i = \bar{\mathbf{A}}_i&h_{i-1} + \bar{\mathbf{B}}_ix'_i\mathrm{,}~~~y'_i = \mathbf{C}_i h'_i + \mathbf{D}x'_i\mathrm{,}~~~y_i = y'_i \odot z_i
\end{align}
where $\sigma$ denotes the SiLU activation, and $\odot$ denotes element-wise multiply.

\vspace{-3mm}
\subsection{Overall architecture of \ourmethod}
\label{subsec:architecture}

In Figure~\ref{fig:pipeline}, we present the model architecture of \ourmethod. Our model is divided into three main components: (1) a convolutional tokenizer that transforms an input 2D image into visual tokens, (2) the main network with a series of $L$ identical \ourmethod~blocks to learn visual representations, and (3) a task-specific head for downstream applications.

In more detail, the tokenizer will downsample the input image $I \in \mathbb{R}^{H_I\times W_I \times 3}$ into a list of visual tokens $x \in \mathbb{R}^{H \times W \times C}$, where $C$ is the channel number. We set the default down-sampling factor to 16, following ViT~\cite{ViT_dosovitskiy2021an}. After combining the initial visual tokens with positional embeddings~\cite{Transformer_NIPS2017_Vaswani} for retaining spatial information, the tokens undergo a series of transformations through the $L$ \ourmethod~blocks, which are designed to simplify usage by maintaining the input-output shape consistency. The final stage of the architecture involves a task-specific head, which is dependent on the particular downstream application. For instance, in image classification tasks, the image tokens are globally pooled into a vector, which is then fed into a linear classification head to produce the final output.

\ourmethod~distinguishes itself from existing vision transformers~\cite{ViT_dosovitskiy2021an,Liu2024Feb_swin_umamba} and concurrent vision Mamba~\cite{liu2024vmamba,zhu2024vision} architectures in several key aspects. Firstly, it does not use any special tokens, such as the commonly used \texttt{CLS} token. Secondly, in contrast to approaches that adopt a hierarchical structure to manage feature resolution~\cite{pvtv2,mvitv2,Swin_Liu_2021tq}, Instead, \ourmethod~maintains a constant feature resolution across all blocks. This design choice considers the recent progress made in various visual foundation models~\cite{sam_model,clip,dinov2} where the plain non-hierarchical ViT is used rather than its hierarchical counterparts.

\vspace{-3mm}
\subsection{\ourmethod~Block}
\label{subsec:simba-block}

The overall architecture comprises several identical \ourmethod~blocks, forming the backbone for learning high-quality visual features. We present the structure of the \ourmethod~block in Figure~\ref{fig:pipeline}, in which we make several key adjustments to the original Mamba block to fully exploit the two-dimensional nature of image inputs. This adaptation is crucial for effectively transitioning from the inherently 1D processing paradigm of language models to the 2D domain of images.
To this end, we introduce two novel techniques: (1) \textit{Continuous 2D Scanning} and (2) \textit{Direction-Aware Updating}. The first technique ensures that each visual token is always adjacent to the previous scanned token. By doing so, it mitigates positional bias and encourages a more uniform understanding of the image space, enhancing the model's ability to learn from visual inputs. The second technique explicitly embeds the 2D relative positional information into the selective scanning process, which allows the model to better interpret the positional context of flattened visual tokens. \\

\begin{figure*}[!t]
    \centering
    \includegraphics[width=0.8\textwidth]{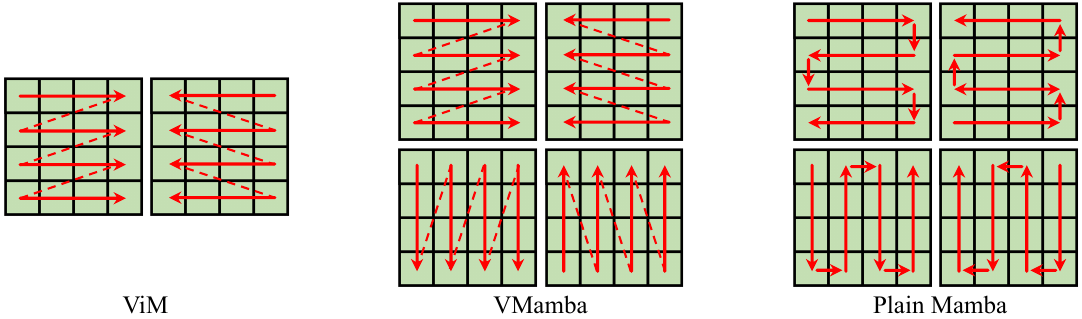}
\vspace{2mm}
    \captionsetup{font={footnotesize}}
\caption{Comparison between our Continuous 2D Scanning and the selective scan orders in ViM~\cite{zhu2024vision} and VMamba~\cite{liu2024vmamba}. Our method makes sure that every scanned visual token is spatially adjacent to its predecessor, avoiding potential spatial and semantic discontinuity. }
\label{fig:continuous-scanning}
\vspace{-5mm}
\end{figure*}

\noindent\textbf{Continuous 2D Scanning.}
The selective scan mechanism is inherently designed for sequential data, such as text. Adapting this mechanism for 2D image data requires flattening the 2D viusal tokens into a 1D sequence to apply the State Space Model (SSM) updating rule. Prior research, e.g., VisionMamba~\cite{zhu2024vision} and VMamba~\cite{liu2024vmamba}, has demonstrated the efficacy of using multiple scanning orders to enhance model performance --- such as both row-wise and column-wise scans in multiple directions. However, as shown in Figure~\ref{fig:continuous-scanning} (a) and (b), in these approaches, each scanning order can only cover one type of 2D direction, e.g., left to right, causing spatial discontinuity when moving to a new row (or column). Moreover, as the parameter $\mathbf{A}$ in Equation~\ref{eq:ssm-output} serves as a decaying term, such spatial discontinuity can also cause adjacent tokens to be decayed to different degrees, compounding the semantic discontinuity and resulting in potential performance drop.

\setlength\intextsep{0pt}
\begin{table}[!t]
    \centering
    \caption{\footnotesize {\ourmethod~ variants. FLOPs are measured using input size 224$\times$224.}}
    \vspace{3mm}
    \begin{adjustbox}{max width=0.4\linewidth}
    \begin{tabular}{l@{\hskip 15pt}c@{\hskip 10pt}c@{\hskip 10pt}r@{\hskip 10pt}r}
        \toprule
        Model & Depth & Channels & Params & FLOPs  \\
        \midrule
        \ourmethod-L1 & 24 & 192 & 7.3M & 3.0G   \\
        \ourmethod-L2 & 24 & 384 & 25.7M & 8.1G   \\
        \ourmethod-L3 & 36 & 448 & 50.5M & 14.4G   \\
        \bottomrule
    \end{tabular}
    \end{adjustbox}
    \label{table:architecture}
    \vspace{-6mm}
\end{table}

Our \textit{Continuous 2D Scanning} addresses this challenge by ensuring a scanned visual token is always adjacent (in the 2D space) to the previously scanned token. As shown in Figure~\ref{fig:continuous-scanning} (c), in our approach, the visual tokens are also scanned in four distinct orders. When reaching the end of a row (or column), the next scanned token will be its adjacent, \textit{not the opposite}, token in the next column (or row). Then, the scanning continues with a reversed direction until it reaches the final visual token of the image. As a consequence, our method preserves spatial and semantic continuity and avoids potential information loss when scanning non-adjacent tokens. Furthermore, in practice the model usually takes input images of the same size, meaning our method can be easily implemented and efficiently run by pre-computing the permutation indexes. \\

\vspace{-2mm}

\noindent\textbf{Direction-Aware Updating.}
As shown in Equation~\ref{eq:ssm-output}, the contribution of a token $x_i$ to the hidden state $h_i$ in the selective scan is determined by the parameter $\bar{\mathbf{B}}_i$, derived from $x_i$ itself. In language models, the sequential order naturally dictates the positional relationship between tokens, allowing the model to "\textit{remember}" their relative positions. However, in our Continuous 2D Scanning, the current token can be in one of four possible directions relative to its predecessor. This challenges the model's ability to discern the precise spatial relationship between consecutive tokens based on $\mathbf{B}_i$ alone. Our \textit{Direction-Aware Updating} is therefore proposed to address this challenge. Drawing inspiration from the relative positional encoding mechanisms in vision transformers~\cite{ViT_dosovitskiy2021an}, we employ a set of learnable parameters $\{\mathbf{\Theta}_k \in \mathbb{R}^{m\times 1}\}$ (for $k=1,2,...,5$), representing the four cardinal directions plus a special \texttt{BEGIN} direction for the initial token. These parameters are summed with the data-dependent $\mathbf{B}_i$ to enrich the selective scan process with directional information. Specifically, with $x_i$ and $z_i$ following Equation~\ref{eq:ssm-output}, our \textit{Direction-Aware Updating} is formulated as follows:
\begin{align}
  &h'_{k,i} = \bar{\mathbf{A}}_{i} h_{k,i-1} + (\bar{\mathbf{B}}_i + \bar{\mathbf{\Theta}}_{k,i})x_i\\
  y'_i =& \sum_{k=1}^{4}\bigl(\mathbf{C}_i h'_{k,i} + \mathbf{D}x_i\bigr)\mathrm{,}~~~y_i = y'_i \odot z_i
\end{align} where $k$ spans the four distinct scanning directions introduced by our \textit{Continuous 2D Scanning}. Alternatively, for the initial token of each scan, we instead add the final $\bar{\mathbf{\Theta}}_{k=5}$ vector.
The term $\bar{\mathbf{\Theta}}_{k,i}$ represents the discretized  $\mathbf{\Theta}_{k,i}$ using $\mathbf{\Delta}_i$.

\vspace{-2mm}
\subsection{Model Variants of \ourmethod}
\label{subsec:simba-variants}

As shown in Table~\ref{table:architecture},  we present three different model variants of \ourmethod. Specifically, from \ourmethod-L1 to \ourmethod-L2, we scale the model width, i.e., feature channel numbers, and keep the model depth to 24. From \ourmethod-L2 to \ourmethod-L3, we scale both model width and depth. The FLOPs are measured using 224$\times$224 inputs, and we follow the official Mamba codebase to compute the FLOPs of the selective scan process. 

\vspace{-0.2in}
\section{Experiments}

\setlength\intextsep{0pt}
\begin{wraptable}{R}{0.5\textwidth}
\centering
\caption{\footnotesize Comparison between \ourmethod~and other models on ImageNet-1K. (\textcolor{red}{$^*$} denotes best epoch result.)}
\vspace{4mm}
\label{table:in1kSmall-big-table}
\begin{adjustbox}{max width=0.5\textwidth}
\begin{tabular}{lcrrr}
\hline
Model   & Hierarchical & Params & FLOPs  & Top-1  \\
\hline
\multicolumn{4}{l}{CNN} \\
\hline
ResNeXt101-32$\times$4~\cite{xie2017aggregated} & \checkmark &  44M &  8.0G &  78.6  \\
ResNeXt101-32$\times$8~\cite{xie2017aggregated} &  \checkmark & 88M &  16.5G &  79.3  \\
RegNetY-4G~\cite{radosavovic2020designing} & \checkmark &21M &  4.0G & 80.0 \\
RegNetY-8G~\cite{radosavovic2020designing} & \checkmark &39M & 8.0G & 81.7\\
ConvNeXt-T~\cite{Liu2022Jan_convnext} & \checkmark &  29M & 4.5G &  82.1 \\
ConvNeXt-S~\cite{Liu2022Jan_convnext} & \checkmark &  50M & 8.7G &  83.1 \\
\hline 
\multicolumn{4}{l}{Transformer} \\
\hline
DeiT-Tiny~\cite{DeiT_touvron2020} & \xmark &   5M&     1.3G &  72.2   \\
DeiT-Small~\cite{DeiT_touvron2020} &  \xmark &  22M &     4.6G &  79.9  \\
DeiT-Base~\cite{DeiT_touvron2020} &  \xmark &  86M &  16.8G &  81.8 \\
Swin-Tiny~\cite{Swin_Liu_2021tq} &   \checkmark & 29M & 4.5G &  81.3  \\
Swin-Small~\cite{Swin_Liu_2021tq} & \checkmark &   50M & 8.7G &  83.0  \\
PVT-Tiny~\cite{PVT_wang2021}  &\checkmark & 13M & 2G & 75.1 \\
PVT-Small~\cite{PVT_wang2021}  & \checkmark &25M & 3.8G & 79.8 \\
PVT-Medium~\cite{PVT_wang2021}  & \checkmark &44M & 6.7G & 81.2 \\
Focal-Tiny~\cite{focal} & \checkmark &29M & 4.9G & 82.2 \\
Focal-Small~\cite{focal} & \checkmark &51M & 9.1G & 83.5 \\
\hline
\multicolumn{4}{l}{State Space Modeling} \\
\hline
ViM-T~\cite{zhu2024vision} & \xmark &   7M & - &  76.1   \\
ViM-S~\cite{zhu2024vision} & \xmark &   26M & - &  80.5   \\
LocalViM-T~\cite{huang2024localmamba} &  \xmark &  8M & 1.5G &  76.2 \\
LocalViM-S~\cite{huang2024localmamba} &  \xmark &  28M & 4.8G &  81.2 \\
Mamba-ND-T~\cite{li2024mamba_nd} & \xmark &24M & - & 79.2 \\
Mamba-ND-S~\cite{li2024mamba_nd} & \xmark &63M & - & 79.4 \\
S4ND-ViT-B~\cite{nguyen2022s4nd} &\xmark & 89M & - & 80.4 \\
S4ND-ConvNeXt-T~\cite{nguyen2022s4nd} & \checkmark &30M & - & 82.2 \\
VMamba-T~\cite{liu2024vmamba} &  \checkmark &  22M & 5.6G & \textcolor{red}{$^*$}82.2 \\
VMamba-S~\cite{liu2024vmamba} &  \checkmark &  44M & 11.2G &  \textcolor{red}{$^*$}83.5 \\

\hline
\ourmethod-L1 & \xmark &7M &   3.0G & 77.9   \\
\ourmethod-L2 & \xmark &25M &   8.1G & 81.6 \\
\ourmethod-L3 & \xmark &50M &   14.4G & 82.3  \\
\hline
\end{tabular}
\end{adjustbox}

\end{wraptable}

In the main paper, we quantitatively compare \ourmethod~with previously proposed models on four visual recognition tasks: image classification, object detection, instance segmentation, and semantic segmentation. Please refer to our supplementary materials for further ablation studies.

\subsection{Experiment Settings}
\label{sec:suppSettings}

\noindent\textbf{ImageNet Classification.}
We build our codebase following~\cite{yang2023gpvit}, which is a commonly used training recipe. Specifically, for the ImageNet-1k experiments, we train all \ourmethod~models for 300 epochs using AdamW optimizer. Following~\cite{Swin_Liu_2021tq}, we set the batch size to 2048, weight decay to 0.05, and the peak learning rate to 0.002. Cosine learning rate scheduling is used. For data augmentation, we used the commonly used recipe~\cite{DeiT_touvron2020,Swin_Liu_2021tq,cswin,DAViT}, which includes Mixup~\cite{zhang2017mixup}, Cutmix~\cite{yun2019cutmix}, Random erasing~\cite{zhong2020random} and Rand augment~\cite{cubuk2020randaugment}.\\

\noindent\textbf{ADE20K Semantic Segmentation.}
We follow common practice~\cite{Swin_Liu_2021tq,cswin,yang2023gpvit} to use UperNet~\cite{xiao2018unified} as the segmentation network. Unlike XCiT~\cite{xcit}, we do not explicitly resize the constant resolution feature maps into multi-scale. Following~\cite{Swin_Liu_2021tq}, we train all models for 160 iterations with batch size 16 and set the default training image size to 512$\times$512. \\

\noindent\textbf{COCO Object Detection and Instance Segmentation.}
Following~\cite{yang2023gpvit}, we test \ourmethod's ability on COCO object detection and instance segmentation using both the two-stage Mask R-CNN~\cite{MaskRCNN_He_2017_ICCV} and the single-stage RetinaNet~\cite{RetinaNet_Lin_2017_ICCV}. For both models, we report the results of both 1$\times$ schedule. Following~\cite{yang2023gpvit}, we use ViTAdapter~\cite{vitadapter} to compute multi-scale features to fit the FPN network structure. We use the commonly used training settings proposed in~\cite{Swin_Liu_2021tq} to keep a fair comparison.

\subsection{Main Results}
\noindent\textbf{ImageNet-1K Classification.}
In Table~\ref{table:in1kSmall-big-table}, we report the ImageNet-1K experiment results. We compare \ourmethod~with three different kinds of visual feature extractors: CNNs, vision transformers, and SSMs. In addition, the comparison includes both hierarchical and non-hierarchical models.  Firstly, when comparing with SSMs, our model is doing better than the recently proposed Vision Mamba~\cite{zhu2024vision} and Mamba-ND~\cite{li2024mamba_nd}. For example, \ourmethod-L2 achieves a 2.4\% higher accuracy than Mamba-ND-T while they share a similar model size. These results validate \ourmethod's effectiveness as a non-hierarchical SSM. Secondly, when compared with CNNs and transformers, our model achieves better performance than the non-hierarchical counterparts. For example, \ourmethod-L2 achieves 1.7\% better accuracy with DeiT-Small. Moreover, \ourmethod~also achieves similar performance when compared with hierarchical models. For example, when the model size is around 25M, our model achieves 0.3\% better accuracy than Swin-Tiny, validating \ourmethod's ability as a general feature extractor. On the other hand, the hierarchical VMamba~\cite{liu2024vmamba}, together with other hierarchical transformers, do achieve a better accuracy than ours. As we explained in Section~\ref{sec:intro}, hierarchical models tend to perform better than non-hierarchical ones in visual recognition. As the main motivation of our work is to develop a simple Mamba architecture, a bit inferior ImageNet accuracy is acceptable. \\

\setlength\intextsep{0pt}
\begin{wraptable}{R}{0.5\textwidth}
\centering
\caption{\footnotesize ADE20K semantic segmentation using UperNet. The FLOPs are computed using input size 512$\times$2048.}%
\label{table:adeLarge}
\vspace{4mm}
\begin{adjustbox}{max width=.5\textwidth}
\begin{tabular}{lc@{\hskip 10pt}r@{\hskip 10pt}r@{\hskip 10pt}|@{\hskip 10pt}c@{\hskip 10pt}}
\hline
Backbone & Hierarchical & Params &  FLOPs & mIoU  \\
\hline
\multicolumn{4}{l}{CNN} \\
\hline
ResNet-50~\cite{resnet} & \checkmark & 67M & 953G & 42.1 \\
ResNet-101~\cite{resnet} & \checkmark & 85M & 1030G & 44.0 \\
ConvNeXt-T~\cite{Liu2022Jan_convnext} & \checkmark & 60M & 939G & 46.7 \\ 
\hline
\multicolumn{4}{l}{Transformer} \\
\hline
DeiT-S+MLN~\cite{DeiT_touvron2020} &    \xmark & 58M &    1217G & 43.8 \\
DeiT-B+MLN~\cite{DeiT_touvron2020} &\xmark & 144M & 2007G & 45.5 \\ 
XCiT-T12/8~\cite{xcit} & \xmark &34M & - & 43.5 \\ 
XCiT-S12/8~\cite{xcit} & \xmark &52M & 1237G & 46.6 \\ 
XCiT-S24/8~\cite{xcit} & \xmark &74M & 1587G &  48.1 \\ 
Swin-Tiny~\cite{Swin_Liu_2021tq} & \checkmark &   60M &    945G & 44.5 \\
Swin-Small~\cite{Swin_Liu_2021tq} &  \checkmark &  81M &    1038G &  47.6 \\
Focal-Tiny~\cite{focal} & \checkmark &62M &    998G &  45.8 \\
Focal-Small~\cite{focal} & \checkmark &85M &    1130G &  48.0 \\

Twins-SVT-Small~\cite{Twins_Chu_2021to}  & \checkmark &54M &    912G &  46.2 \\
Twins-SVT-Small~\cite{Twins_Chu_2021to}  & \checkmark &88M &    1044G &  47.7 \\

\hline
\multicolumn{4}{l}{State Space Modeling} \\
\hline
ViM-T~\cite{zhu2024vision} & \xmark &13M & - &  41.0 \\
ViM-S~\cite{zhu2024vision} & \xmark & 46M & - &  44.9 \\
LocalVim-T~\cite{huang2024localmamba} & \xmark & 36M & 181G &  43.4 \\
LocalVim-S~\cite{huang2024localmamba} & \xmark & 58M & 297G &  46.4 \\
VMamba-T~\cite{liu2024vmamba} & \checkmark & 55M & 964G & 47.3 \\
VMamba-S~\cite{liu2024vmamba} & \checkmark & 76M & 1081G & 49.5 \\
\hline
\ourmethod-L1 & \xmark &35M & 174G & 44.1  \\
\ourmethod-L2 & \xmark &55M & 285G & 46.8  \\
\ourmethod-L3 & \xmark &81M & 419G & 49.1  \\
\hline
\end{tabular}
\end{adjustbox}

\end{wraptable}

\noindent\textbf{ADE20K Semantic Segmentation}
We report our model's ADE20K semantic segmentation performance in Table~\ref{table:adeLarge}. Similar to the ImageNet-1k and COCO experiments, here the competing models include both hierarchical and non-hierarchical backbones in three types of visual feature extractors. The results again suggest that \ourmethod~achieves the best performance among the non-hierarchical models. For example, with similar parameter amounts, \ourmethod-L2 outperforms the high-resolution (patch size of 8) XCiT-S12/8 model~\cite{xcit} with a much lower computation cost. Moreover, \ourmethod-L2 also outperforms the hierarchical Swin-Transformer-Tiny~\cite{Swin_Liu_2021tq}, achieving better mIoU while having a lower model size and FLOPs. At the same time, \ourmethod~is also doing better than the concurrent Vision Mamba~\cite{zhu2024vision}. For instance, \ourmethod-L2 achieves a 1.9 higher mIoU than ViM-S. This result verifies our model's effectiveness in extracting fine-grained visual features, which is essential for the pixel-wise semantic segmentation task. \\

\begin{table}[!t]
\centering
\caption{\footnotesize Mask R-CNN object detection and instance segmentation on MS COCO \textit{mini-val} using 1$\times$ schedule. We use ViTAdapter~\cite{vitadapter} to compute multi-scale features. FLOPs are computed using input size  1280$\times$800. }
\label{table:maskrcnnSmall}
\vspace{4mm}
\begin{adjustbox}{max width=.87\textwidth}
\begin{tabular}{lcr@{\hskip 5pt}r@{\hskip 5pt}|ccc|ccc}
\hline
Backbone &  Hierarchical & Params &  FLOPs & $AP^{bb}$  & $AP^{bb}_{50}$ & $AP^{bb}_{75}$  & $AP^{mk}$ & $AP^{mk}_{50}$  & $AP^{mk}_{75}$ \\
\hline
\multicolumn{10}{l}{CNN} \\
\hline
ResNeXt101-32x4d~\cite{xie2017aggregated} & \checkmark & 63M & 340G & 41.9 & - & - & 37.5 & - & - \\
ResNeXt101-64x4d~\cite{xie2017aggregated} & \checkmark &102M & 493G & 42.8 & - & - & 38.4 & - & -  \\
\hline
\multicolumn{10}{l}{Transformer} \\
\hline
ViT-Adapter-T~\cite{vitadapter} & \xmark &29M & 349G & 41.1 & 62.5 & 44.3 & 37.5 & 59.7 & 39.9  \\
ViT-Adapter-S~\cite{vitadapter} &\xmark & 49M & 463G & 44.7 & 65.8 &  48.3 &  39.9 &  62.5 &  42.8 \\
ViT-Adapter-B~\cite{vitadapter} &\xmark & 131M & 838G & 47.0 & 68.2 & 51.4 &  41.8 &  65.1 &  44.9 \\

PVT-Small~\cite{PVT_wang2021} & \checkmark &44M & -& 40.4 & 62.9 & 43.8 & 37.8 & 60.1 & 40.3 \\
PVT-Medium~\cite{PVT_wang2021} & \checkmark &64M & - & 42.0 & 64.4 & 45.6 & 39.0 & 61.6 & 42.1 \\
PVT-Large~\cite{PVT_wang2021} & \checkmark &81M & - & 42.9 & 65.0 & 46.6 & 39.5 & 61.9 & 42.5 \\
Swin-Tiny~\cite{Swin_Liu_2021tq} & \checkmark &48M & 264G & 42.2 & - & - & 39.1 & - & - \\
Swin-Small~\cite{Swin_Liu_2021tq} & \checkmark &69M & 354G & 44.8 & - & - & 40.9 & - & - \\

ViL-Tiny~\cite{ViL_Zhang_2021tu} &\checkmark & 26M & 145G & 41.4 & 63.5 &  45.0  & 38.1 & 60.3 & 40.8 \\
ViL-Small~\cite{ViL_Zhang_2021tu} & \checkmark& 45M & 218G & 44.9 & 67.1 & 49.3 & 41.0 & 64.2 & 44.1 \\
ViL-Medium~\cite{ViL_Zhang_2021tu} &\checkmark & 60M & 293G & 47.6 & 69.8 & 52.1 &  43.0  & 66.9 &  46.6 \\

\hline
\multicolumn{10}{l}{State Space Modeling} \\
\hline

EfficientVMamba-T~\cite{pei2024efficientvmamba} &\checkmark & 11M & 60G & 35.6 & 57.7 & 38.0 & 33.2 & 54.4 & 35.1 \\
EfficientVMamba-S~\cite{pei2024efficientvmamba} &\checkmark & 31M & 197G & 39.3 & 61.8 & 42.6 & 36.7 & 58.9 & 39.2 \\
EfficientVMamba-B~\cite{pei2024efficientvmamba} &\checkmark & 53M & 252G & 43.7 & 66.2 & 47.9 & 40.2 & 63.3 & 42.9 \\

VMamba-T~\cite{liu2024vmamba} & \checkmark &42M & 262G & 46.5 & 68.5 & 50.7 & 42.1 & 65.5 & 45.3 \\
VMamba-S~\cite{liu2024vmamba} & \checkmark &64M & 357G & 48.2 & 69.7 & 52.5 & 43.0 & 66.6 & 46.4  \\

\hline
\ourmethod-Adapter-L1 & \xmark &31M & 388G & 44.1 & 64.8 & 47.9 & 39.1 & 61.6 & 41.9 \\
\ourmethod-Adapter-L2 & \xmark &53M & 542G & 46.0 & 66.9 & 50.1 & 40.6 & 63.8 & 43.6 \\
\ourmethod-Adapter-L3 & \xmark &79M & 696G & 46.8 & 68.0 & 51.1 & 41.2 & 64.7 & 43.9 \\
\hline
\end{tabular}
\end{adjustbox}
\vspace{-2mm}

\end{table}

\noindent\textbf{COCO Object Detection and Instance Segmentation.}
We report the results of Mask R-CNN object detection and instance segmentation in Table~\ref{table:maskrcnnSmall}. With similar FLOPs and many fewer parameters, \ourmethod-L1 achieves 44.1 AP$^{bb}$ and 39.1 AP$^{mk}$ when using 1$\times$ training schedule, while Swin-Small achieves 44.8 AP$^{bb}$ and 40.9 AP$^{mk}$. We also observe that hierarchical models tend to work better than non-hierarchical models. Although our model achieves lower performance than some hierarchical models, e.g., the concurrent VMamba~\cite{liu2024vmamba}, \ourmethod~achieves the best performance among its non-hierarchical counterparts. For instance, when using 1$\times$ training schedule, \ourmethod~achieves 3.1 higher AP$^{bb}$ and 1.6 higher AP$^{mk}$ than DeiT-T when they are both equipped with the ViTAdapter~\cite{vitadapter}.  These results demonstrate that \ourmethod~is able to extract good local features, which is important to the object-level tasks like instance segmentation. On the other hand, we also admit that \ourmethod~is performing worse than the hierarchical VMamba~\cite{liu2024vmamba}. We attribute such inferiority to the multi-resolution architecture of FPN-based~\cite{lin2017feature} Mask R-CNN, which is more naturally suitable to the hierarchical designs.

\vspace{-2mm}
\section{Conclusion}

We present~\ourmethod, a plain SSM-based model for visual recognition. Our model is conceptually simple because it uses no special tokens or hierarchical structure, making it a perfect counterpart to the widely used plain vision transformer. The results show that \ourmethod~achieves superior performance to previous non-hierarchical models, including the concurrent SSM-based models, and can perform on par with the high-performing hierarchical models. We hope our model can serve as a baseline for future research in this area. 

\section*{Acknowledgements}
Funding for this research is provided in part by a studentship from the School of Engineering at the University of Edinburgh, the SENSE - Centre for Satellite Data in Environmental Science CDT, and an EPSRC New Investigator Award (EP/X020703/1).

\clearpage

\clearpage

\bibliography{main}

\begin{thebibliography}{110}
\providecommand{\natexlab}[1]{#1}
\providecommand{\url}[1]{\texttt{#1}}
\expandafter\ifx\csname urlstyle\endcsname\relax
  \providecommand{\doi}[1]{doi: #1}\else
  \providecommand{\doi}{doi: \begingroup \urlstyle{rm}\Url}\fi

\bibitem[Ali et~al.(2021)Ali, Touvron, Caron, Bojanowski, Douze, Joulin, Laptev, Neverova, Synnaeve, Verbeek, et~al.]{xcit}
Alaaeldin Ali, Hugo Touvron, Mathilde Caron, Piotr Bojanowski, Matthijs Douze, Armand Joulin, Ivan Laptev, Natalia Neverova, Gabriel Synnaeve, Jakob Verbeek, et~al.
\newblock Xcit: Cross-covariance image transformers.
\newblock \emph{Advances in neural information processing systems}, 34:\penalty0 20014--20027, 2021.

\bibitem[Baron et~al.(2023)Baron, Zimerman, and Wolf]{Baron2023Jun_2d_ssm}
Ethan Baron, Itamar Zimerman, and Lior Wolf.
\newblock {2-D ssm: A general spatial layer for visual transformers}.
\newblock \emph{arXiv preprint arXiv:2306.06635}, 2023.

\bibitem[Bay et~al.(2006)Bay, Tuytelaars, and Van~Gool]{Bay2006_surf}
Herbert Bay, Tinne Tuytelaars, and Luc Van~Gool.
\newblock {SURF: Speeded Up Robust Features}.
\newblock In \emph{ECCV}, 2006.

\bibitem[Bello(2021)]{lambdanetworks_bello2021}
Irwan Bello.
\newblock Lambdanetworks: Modeling long-range interactions without attention.
\newblock In \emph{International Conference on Learning Representations}, 2021.
\newblock URL \url{https://openreview.net/forum?id=xTJEN-ggl1b}.

\bibitem[Beyer et~al.(2020)Beyer, H{\'e}naff, Kolesnikov, Zhai, and Oord]{donewithimagenet}
Lucas Beyer, Olivier~J H{\'e}naff, Alexander Kolesnikov, Xiaohua Zhai, and A{\"a}ron van~den Oord.
\newblock Are we done with imagenet?
\newblock \emph{arXiv preprint arXiv:2006.07159}, 2020.

\bibitem[Beyer et~al.(2023)Beyer, Izmailov, Kolesnikov, Caron, Kornblith, Zhai, Minderer, Tschannen, Alabdulmohsin, and Pavetic]{Beyer2022Dec_flexivit}
Lucas Beyer, Pavel Izmailov, Alexander Kolesnikov, Mathilde Caron, Simon Kornblith, Xiaohua Zhai, Matthias Minderer, Michael Tschannen, Ibrahim Alabdulmohsin, and Filip Pavetic.
\newblock {FlexiViT: One Model for All Patch Sizes}.
\newblock In \emph{CVPR}, 2023.

\bibitem[Brown et~al.(2020)Brown, Mann, Ryder, Subbiah, Kaplan, Dhariwal, Neelakantan, Shyam, Sastry, Askell, Agarwal, Herbert-Voss, Krueger, Henighan, Child, Ramesh, Ziegler, Wu, Winter, Hesse, Chen, Sigler, Litwin, Gray, Chess, Clark, Berner, McCandlish, Radford, Sutskever, and Amodei]{gpt3}
Tom Brown, Benjamin Mann, Nick Ryder, Melanie Subbiah, Jared~D Kaplan, Prafulla Dhariwal, Arvind Neelakantan, Pranav Shyam, Girish Sastry, Amanda Askell, Sandhini Agarwal, Ariel Herbert-Voss, Gretchen Krueger, Tom Henighan, Rewon Child, Aditya Ramesh, Daniel Ziegler, Jeffrey Wu, Clemens Winter, Chris Hesse, Mark Chen, Eric Sigler, Mateusz Litwin, Scott Gray, Benjamin Chess, Jack Clark, Christopher Berner, Sam McCandlish, Alec Radford, Ilya Sutskever, and Dario Amodei.
\newblock Language models are few-shot learners.
\newblock In H.~Larochelle, M.~Ranzato, R.~Hadsell, M.F. Balcan, and H.~Lin, editors, \emph{NeurIPS}, 2020.

\bibitem[Chen et~al.(2022{\natexlab{a}})Chen, Panda, and Fan]{chen2022regionvit}
Chun-Fu Chen, Rameswar Panda, and Quanfu Fan.
\newblock Regionvit: Regional-to-local attention for vision transformers.
\newblock In \emph{International Conference on Learning Representations}, 2022{\natexlab{a}}.

\bibitem[Chen et~al.(2019)Chen, Fan, Xu, Yan, Kalantidis, Rohrbach, Yan, and Feng]{chen2019drop}
Yunpeng Chen, Haoqi Fan, Bing Xu, Zhicheng Yan, Yannis Kalantidis, Marcus Rohrbach, Shuicheng Yan, and Jiashi Feng.
\newblock Drop an octave: Reducing spatial redundancy in convolutional neural networks with octave convolution.
\newblock In \emph{Proceedings of the IEEE/CVF International Conference on Computer Vision}, pages 3435--3444, 2019.

\bibitem[Chen et~al.(2022{\natexlab{b}})Chen, Duan, Wang, He, Lu, Dai, and Qiao]{vitadapter}
Zhe Chen, Yuchen Duan, Wenhai Wang, Junjun He, Tong Lu, Jifeng Dai, and Yu~Qiao.
\newblock Vision transformer adapter for dense predictions.
\newblock \emph{arXiv preprint arXiv:2205.08534}, 2022{\natexlab{b}}.

\bibitem[Cheng et~al.(2022)Cheng, Misra, Schwing, Kirillov, and Girdhar]{cheng2021mask2former}
Bowen Cheng, Ishan Misra, Alexander~G. Schwing, Alexander Kirillov, and Rohit Girdhar.
\newblock Masked-attention mask transformer for universal image segmentation.
\newblock In \emph{Proceedings of the IEEE/CVF Conference on Computer Vision and Pattern Recognition}, June 2022.

\bibitem[Chollet(2016)]{Chollet2016Oct}
Fran{\ifmmode\mbox{\c{c}}\else\c{c}\fi}ois Chollet.
\newblock {Xception: Deep Learning with Depthwise Separable Convolutions}.
\newblock In \emph{CVPR}, 2016.

\bibitem[Chu et~al.(2021)Chu, Tian, Wang, Zhang, Ren, Wei, Xia, and Shen]{Twins_Chu_2021to}
Xiangxiang Chu, Zhi Tian, Yuqing Wang, Bo~Zhang, Haibing Ren, Xiaolin Wei, Huaxia Xia, and Chunhua Shen.
\newblock {Twins: Revisiting Spatial Attention Design in Vision Transformers}.
\newblock \emph{arXiv.org}, April 2021.

\bibitem[Cubuk et~al.(2020)Cubuk, Zoph, Shlens, and Le]{cubuk2020randaugment}
Ekin~D Cubuk, Barret Zoph, Jonathon Shlens, and Quoc~V Le.
\newblock Randaugment: Practical automated data augmentation with a reduced search space.
\newblock In \emph{Proceedings of the IEEE/CVF conference on computer vision and pattern recognition workshops}, pages 702--703, 2020.

\bibitem[Dao(2024)]{flash_attention2}
Tri Dao.
\newblock Flash{A}ttention-2: Faster attention with better parallelism and work partitioning.
\newblock In \emph{ICLR}, 2024.

\bibitem[Dao et~al.(2022{\natexlab{a}})Dao, Fu, Ermon, Rudra, and R{\'e}]{dao2022flashattention}
Tri Dao, Dan Fu, Stefano Ermon, Atri Rudra, and Christopher R{\'e}.
\newblock Flashattention: Fast and memory-efficient exact attention with io-awareness.
\newblock \emph{Advances in Neural Information Processing Systems}, 35:\penalty0 16344--16359, 2022{\natexlab{a}}.

\bibitem[Dao et~al.(2022{\natexlab{b}})Dao, Fu, Ermon, Rudra, and R{\'e}]{flash_attention}
Tri Dao, Daniel~Y. Fu, Stefano Ermon, Atri Rudra, and Christopher R{\'e}.
\newblock Flash{A}ttention: Fast and memory-efficient exact attention with {IO}-awareness.
\newblock In \emph{NeurIPS}, 2022{\natexlab{b}}.

\bibitem[Darcet et~al.(2023)Darcet, Oquab, Mairal, and Bojanowski]{vits_need_registers}
Timothée Darcet, Maxime Oquab, Julien Mairal, and Piotr Bojanowski.
\newblock Vision transformers need registers, 2023.

\bibitem[Deng et~al.(2009)Deng, Dong, Socher, Li, Li, and Fei-Fei]{imagenet_deng2009}
Jia Deng, Wei Dong, Richard Socher, Li-Jia Li, Kai Li, and Li~Fei-Fei.
\newblock Imagenet: A large-scale hierarchical image database.
\newblock In \emph{2009 IEEE conference on computer vision and pattern recognition}, pages 248--255. Ieee, 2009.

\bibitem[Devlin et~al.(2019)Devlin, Chang, Lee, and Toutanova]{bert_devlin-etal-2019-bert}
Jacob Devlin, Ming-Wei Chang, Kenton Lee, and Kristina Toutanova.
\newblock {BERT}: Pre-training of deep bidirectional transformers for language understanding.
\newblock In \emph{Proceedings of the 2019 Conference of the North {A}merican Chapter of the Association for Computational Linguistics: Human Language Technologies, Volume 1 (Long and Short Papers)}, pages 4171--4186, Minneapolis, Minnesota, June 2019. Association for Computational Linguistics.
\newblock \doi{10.18653/v1/N19-1423}.
\newblock URL \url{https://www.aclweb.org/anthology/N19-1423}.

\bibitem[Ding et~al.(2022)Ding, Xiao, Codella, Luo, Wang, and Yuan]{DAViT}
Mingyu Ding, Bin Xiao, Noel Codella, Ping Luo, Jingdong Wang, and Lu~Yuan.
\newblock Davit: Dual attention vision transformers.
\newblock In \emph{Proceedings of the European conference on computer vision}, 2022.

\bibitem[Dong et~al.(2022)Dong, Bao, Chen, Zhang, Yu, Yuan, Chen, and Guo]{cswin}
Xiaoyi Dong, Jianmin Bao, Dongdong Chen, Weiming Zhang, Nenghai Yu, Lu~Yuan, Dong Chen, and Baining Guo.
\newblock Cswin transformer: A general vision transformer backbone with cross-shaped windows.
\newblock In \emph{Proceedings of the IEEE/CVF Conference on Computer Vision and Pattern Recognition}, pages 12124--12134, June 2022.

\bibitem[Dosovitskiy et~al.(2021)Dosovitskiy, Beyer, Kolesnikov, Weissenborn, Zhai, Unterthiner, Dehghani, Minderer, Heigold, Gelly, Uszkoreit, and Houlsby]{ViT_dosovitskiy2021an}
Alexey Dosovitskiy, Lucas Beyer, Alexander Kolesnikov, Dirk Weissenborn, Xiaohua Zhai, Thomas Unterthiner, Mostafa Dehghani, Matthias Minderer, Georg Heigold, Sylvain Gelly, Jakob Uszkoreit, and Neil Houlsby.
\newblock An image is worth 16x16 words: Transformers for image recognition at scale.
\newblock In \emph{International Conference on Learning Representations}, 2021.
\newblock URL \url{https://openreview.net/forum?id=YicbFdNTTy}.

\bibitem[d’Ascoli et~al.(2021)d’Ascoli, Touvron, Leavitt, Morcos, Biroli, and Sagun]{ConViTdAscoli_2021vz}
St{\'e}phane d’Ascoli, Hugo Touvron, Matthew~L Leavitt, Ari~S Morcos, Giulio Biroli, and Levent Sagun.
\newblock Convit: Improving vision transformers with soft convolutional inductive biases.
\newblock In \emph{International Conference on Machine Learning}. PMLR, 2021.

\bibitem[Ericsson et~al.(2021)Ericsson, Gouk, and Hospedales]{ericsson2021well}
Linus Ericsson, Henry Gouk, and Timothy~M Hospedales.
\newblock How well do self-supervised models transfer?
\newblock In \emph{CVPR}, 2021.

\bibitem[et~al(2022)]{dalle2}
Aditya~Ramesh et~al.
\newblock Hierarchical text-conditional image generation with clip latents, 2022.

\bibitem[Fan et~al.(2021)Fan, Xiong, Mangalam, Li, Yan, Malik, and Feichtenhofer]{fan2021multiscale}
Haoqi Fan, Bo~Xiong, Karttikeya Mangalam, Yanghao Li, Zhicheng Yan, Jitendra Malik, and Christoph Feichtenhofer.
\newblock Multiscale vision transformers.
\newblock In \emph{Proceedings of the IEEE/CVF International Conference on Computer Vision}, pages 6824--6835, 2021.

\bibitem[Fei et~al.(2024)Fei, Fan, Yu, and Huang]{Fei2024Feb_mamba_diffusion}
Zhengcong Fei, Mingyuan Fan, Changqian Yu, and Junshi Huang.
\newblock Scalable diffusion models with state space backbone.
\newblock \emph{arXiv preprint arXiv:2402.05608}, 2024.

\bibitem[Fu et~al.(2023)Fu, Dao, Saab, Thomas, Rudra, and R{\ifmmode\acute{e}\else\'{e}\fi}]{Fu2022DecHippo}
Daniel~Y. Fu, Tri Dao, Khaled~K. Saab, Armin~W. Thomas, Atri Rudra, and Christopher R{\ifmmode\acute{e}\else\'{e}\fi}.
\newblock {Hungry Hungry Hippos: Towards Language Modeling with State Space Models}.
\newblock In \emph{ICLR}, 2023.

\bibitem[Graham et~al.(2021)Graham, El-Nouby, Touvron, Stock, Joulin, J\'egou, and Douze]{LeViT_BenGraham_2021vh}
Benjamin Graham, Alaaeldin El-Nouby, Hugo Touvron, Pierre Stock, Armand Joulin, Herv\'e J\'egou, and Matthijs Douze.
\newblock Levit: A vision transformer in convnet's clothing for faster inference.
\newblock In \emph{Proceedings of the IEEE/CVF International Conference on Computer Vision}, 2021.

\bibitem[Gu and Dao(2023)]{mamba}
Albert Gu and Tri Dao.
\newblock Mamba: Linear-time sequence modeling with selective state spaces.
\newblock \emph{arXiv preprint arXiv:2312.00752}, 2023.

\bibitem[Gu et~al.(2021)Gu, Johnson, Goel, Saab, Dao, Rudra, and Re]{Gu2021NovLSSL}
Albert Gu, Isys Johnson, Karan Goel, Khaled~Kamal Saab, Tri Dao, Atri Rudra, and Christopher Re.
\newblock {Combining Recurrent, Convolutional, and Continuous-time Models with Linear State Space Layers}.
\newblock In \emph{NeurIPS}, 2021.

\bibitem[Gu et~al.(2022)Gu, Goel, and Re]{Gu2021OctS4}
Albert Gu, Karan Goel, and Christopher Re.
\newblock Efficiently modeling long sequences with structured state spaces.
\newblock In \emph{ICLR}, 2022.

\bibitem[Gu et~al.(2023)Gu, Johnson, Timalsina, Rudra, and Ré]{hippo}
Albert Gu, Isys Johnson, Aman Timalsina, Atri Rudra, and Christopher Ré.
\newblock How to train your hippo: State space models with generalized orthogonal basis projections.
\newblock In \emph{ICLR}, 2023.

\bibitem[Guo et~al.(2024)Guo, Li, Dai, Ouyang, Ren, and Xia]{Guo2024Feb_mambair}
Hang Guo, Jinmin Li, Tao Dai, Zhihao Ouyang, Xudong Ren, and Shu-Tao Xia.
\newblock Mambair: A simple baseline for image restoration with state-space model.
\newblock In \emph{ECCV}, 2024.

\bibitem[Guo et~al.(2022)Guo, Han, Wu, Tang, Chen, Wang, and Xu]{CMT}
Jianyuan Guo, Kai Han, Han Wu, Yehui Tang, Xinghao Chen, Yunhe Wang, and Chang Xu.
\newblock Cmt: Convolutional neural networks meet vision transformers.
\newblock In \emph{Proceedings of the IEEE/CVF Conference on Computer Vision and Pattern Recognition}, pages 12175--12185, June 2022.

\bibitem[Hatamizadeh et~al.(2022)Hatamizadeh, Yin, Kautz, and Molchanov]{gcvit}
Ali Hatamizadeh, Hongxu Yin, Jan Kautz, and Pavlo Molchanov.
\newblock Global context vision transformers.
\newblock \emph{arXiv preprint arXiv:2206.09959}, 2022.

\bibitem[He et~al.(2016)He, Zhang, Ren, and Sun]{resnet}
Kaiming He, Xiangyu Zhang, Shaoqing Ren, and Jian Sun.
\newblock {Deep Residual Learning for Image Recognition}.
\newblock In \emph{The IEEE Conference on Computer Vision and Pattern Recognition}, June 2016.

\bibitem[He et~al.(2017)He, Gkioxari, Dollar, and Girshick]{MaskRCNN_He_2017_ICCV}
Kaiming He, Georgia Gkioxari, Piotr Dollar, and Ross Girshick.
\newblock Mask r-cnn.
\newblock In \emph{Proceedings of the IEEE International Conference on Computer Vision}, Oct 2017.

\bibitem[He et~al.(2024)He, Cao, Yan, Li, Xie, Zhang, and Zhou]{He2024Feb_panmamba}
Xuanhua He, Ke~Cao, Keyu Yan, Rui Li, Chengjun Xie, Jie Zhang, and Man Zhou.
\newblock Pan-mamba: Effective pan-sharpening with state space model.
\newblock \emph{arXiv preprint arXiv:2402.12192}, 2024.

\bibitem[Hou et~al.(2022)Hou, Jiang, Yuan, Cheng, Yan, and Feng]{vision_permutator}
Qibin Hou, Zihang Jiang, Li~Yuan, Ming-Ming Cheng, Shuicheng Yan, and Jiashi Feng.
\newblock Vision permutator: A permutable mlp-like architecture for visual recognition.
\newblock \emph{TPAMI}, 2022.

\bibitem[{Hu} et~al.(2018){Hu}, {Shen}, and {Sun}]{SENet_Hu_2018}
J.~{Hu}, L.~{Shen}, and G.~{Sun}.
\newblock Squeeze-and-excitation networks.
\newblock In \emph{2018 IEEE/CVF Conference on Computer Vision and Pattern Recognition}, pages 7132--7141, 2018.
\newblock \doi{10.1109/CVPR.2018.00745}.

\bibitem[Hua et~al.(2022)Hua, Dai, Liu, and Le]{gated}
Weizhe Hua, Zihang Dai, Hanxiao Liu, and Quoc Le.
\newblock Transformer quality in linear time.
\newblock In \emph{ICML}, 2022.

\bibitem[Huang et~al.(2017)Huang, Liu, van~der Maaten, and Weinberger]{densenet}
Gao Huang, Zhuang Liu, Laurens van~der Maaten, and Kilian~Q. Weinberger.
\newblock Densely connected convolutional networks.
\newblock In \emph{CVPR}, 2017.

\bibitem[Huang et~al.(2024)Huang, Pei, You, Wang, Qian, and Xu]{huang2024localmamba}
Tao Huang, Xiaohuan Pei, Shan You, Fei Wang, Chen Qian, and Chang Xu.
\newblock Localmamba: Visual state space model with windowed selective scan.
\newblock \emph{arXiv preprint arXiv:2403.09338}, 2024.

\bibitem[Kirillov et~al.(2023)Kirillov, Mintun, Ravi, Mao, Rolland, Gustafson, Xiao, Whitehead, Berg, Lo, Doll{\'a}r, and Girshick]{sam_model}
Alexander Kirillov, Eric Mintun, Nikhila Ravi, Hanzi Mao, Chloe Rolland, Laura Gustafson, Tete Xiao, Spencer Whitehead, Alexander~C. Berg, Wan-Yen Lo, Piotr Doll{\'a}r, and Ross Girshick.
\newblock Segment anything.
\newblock \emph{arXiv:2304.02643}, 2023.

\bibitem[Krizhevsky et~al.(2012)Krizhevsky, Sutskever, and Hinton]{alexnet}
Alex Krizhevsky, Ilya Sutskever, and Geoffrey~E Hinton.
\newblock Imagenet classification with deep convolutional neural networks.
\newblock In \emph{NeurIPS}, 2012.

\bibitem[Lee et~al.(2022)Lee, Kim, Willette, and Hwang]{MPViT}
Youngwan Lee, Jonghee Kim, Jeffrey Willette, and Sung~Ju Hwang.
\newblock Mpvit: Multi-path vision transformer for dense prediction.
\newblock In \emph{Proceedings of the IEEE/CVF Conference on Computer Vision and Pattern Recognition}, pages 7287--7296, June 2022.

\bibitem[Li et~al.(2024)Li, Singh, and Grover]{li2024mamba_nd}
Shufan Li, Harkanwar Singh, and Aditya Grover.
\newblock Mamba-nd: Selective state space modeling for multi-dimensional data.
\newblock \emph{arXiv preprint arXiv:2402.05892}, 2024.

\bibitem[Li et~al.(2022{\natexlab{a}})Li, Mao, Girshick, and He]{li2022exploring}
Yanghao Li, Hanzi Mao, Ross Girshick, and Kaiming He.
\newblock Exploring plain vision transformer backbones for object detection.
\newblock In \emph{Proceedings of the IEEE conference on computer vision and pattern recognition}, 2022{\natexlab{a}}.

\bibitem[Li et~al.(2022{\natexlab{b}})Li, Wu, Fan, Mangalam, Xiong, Malik, and Feichtenhofer]{mvitv2}
Yanghao Li, Chao-Yuan Wu, Haoqi Fan, Karttikeya Mangalam, Bo~Xiong, Jitendra Malik, and Christoph Feichtenhofer.
\newblock Mvitv2: Improved multiscale vision transformers for classification and detection.
\newblock In \emph{Proceedings of the IEEE/CVF Conference on Computer Vision and Pattern Recognition}, pages 4804--4814, June 2022{\natexlab{b}}.

\bibitem[Lin et~al.(2014)Lin, Maire, Belongie, Hays, Perona, Ramanan, Doll{\'a}r, and Zitnick]{COCO_Lin_2014vm}
Tsung-Yi Lin, Michael Maire, Serge Belongie, James Hays, Pietro Perona, Deva Ramanan, Piotr Doll{\'a}r, and C~Lawrence Zitnick.
\newblock Microsoft coco: Common objects in context.
\newblock In \emph{ECCV}, 2014.

\bibitem[Lin et~al.(2017{\natexlab{a}})Lin, Doll{\'a}r, Girshick, He, Hariharan, and Belongie]{lin2017feature}
Tsung-Yi Lin, Piotr Doll{\'a}r, Ross Girshick, Kaiming He, Bharath Hariharan, and Serge Belongie.
\newblock Feature pyramid networks for object detection.
\newblock In \emph{Proceedings of the IEEE conference on computer vision and pattern recognition}, pages 2117--2125, 2017{\natexlab{a}}.

\bibitem[Lin et~al.(2017{\natexlab{b}})Lin, Goyal, Girshick, He, and Dollar]{RetinaNet_Lin_2017_ICCV}
Tsung-Yi Lin, Priya Goyal, Ross Girshick, Kaiming He, and Piotr Dollar.
\newblock Focal loss for dense object detection.
\newblock In \emph{Proceedings of the IEEE International Conference on Computer Vision}, Oct 2017{\natexlab{b}}.

\bibitem[Liu et~al.(2023{\natexlab{a}})Liu, Li, Li, and Lee]{improvedllava}
Haotian Liu, Chunyuan Li, Yuheng Li, and Yong~Jae Lee.
\newblock Improved baselines with visual instruction tuning, 2023{\natexlab{a}}.

\bibitem[Liu et~al.(2023{\natexlab{b}})Liu, Li, Wu, and Lee]{llava}
Haotian Liu, Chunyuan Li, Qingyang Wu, and Yong~Jae Lee.
\newblock Visual instruction tuning.
\newblock In \emph{NeurIPS}, 2023{\natexlab{b}}.

\bibitem[Liu et~al.(2024{\natexlab{a}})Liu, Yang, Zhou, Xi, Yu, Yu, Liang, Shi, Zhang, Zheng, et~al.]{Liu2024Feb_swin_umamba}
Jiarun Liu, Hao Yang, Hong-Yu Zhou, Yan Xi, Lequan Yu, Yizhou Yu, Yong Liang, Guangming Shi, Shaoting Zhang, Hairong Zheng, et~al.
\newblock Swin-umamba: Mamba-based unet with imagenet-based pretraining.
\newblock \emph{arXiv preprint arXiv:2402.03302}, 2024{\natexlab{a}}.

\bibitem[Liu et~al.(2024{\natexlab{b}})Liu, Yang, Zhou, Xi, Yu, Yu, Liang, Shi, Zhang, Zheng, et~al.]{liu2024swin}
Jiarun Liu, Hao Yang, Hong-Yu Zhou, Yan Xi, Lequan Yu, Yizhou Yu, Yong Liang, Guangming Shi, Shaoting Zhang, Hairong Zheng, et~al.
\newblock Swin-umamba: Mamba-based unet with imagenet-based pretraining.
\newblock \emph{arXiv preprint arXiv:2402.03302}, 2024{\natexlab{b}}.

\bibitem[Liu et~al.(2024{\natexlab{c}})Liu, Tian, Zhao, Yu, Xie, Wang, Ye, and Liu]{liu2024vmamba}
Yue Liu, Yunjie Tian, Yuzhong Zhao, Hongtian Yu, Lingxi Xie, Yaowei Wang, Qixiang Ye, and Yunfan Liu.
\newblock Vmamba: Visual state space model.
\newblock \emph{arXiv preprint arXiv:2401.10166}, 2024{\natexlab{c}}.

\bibitem[Liu et~al.(2021)Liu, Lin, Cao, Hu, Wei, Zhang, Lin, and Guo]{Swin_Liu_2021tq}
Ze~Liu, Yutong Lin, Yue Cao, Han Hu, Yixuan Wei, Zheng Zhang, Stephen Lin, and Baining Guo.
\newblock {Swin Transformer: Hierarchical Vision Transformer using Shifted Windows}.
\newblock In \emph{Proceedings of the IEEE/CVF International Conference on Computer Vision}, 2021.

\bibitem[Liu et~al.(2022)Liu, Mao, Wu, Feichtenhofer, Darrell, and Xie]{Liu2022Jan_convnext}
Zhuang Liu, Hanzi Mao, Chao-Yuan Wu, Christoph Feichtenhofer, Trevor Darrell, and Saining Xie.
\newblock {A ConvNet for the 2020s}.
\newblock In \emph{CVPR}, 2022.

\bibitem[Lowe(2004)]{lowe2004distinctive_sift}
David~G Lowe.
\newblock Distinctive image features from scale-invariant keypoints.
\newblock In \emph{IJCV}, 2004.

\bibitem[Ma et~al.(2024)Ma, Li, and Wang]{ma2024u}
Jun Ma, Feifei Li, and Bo~Wang.
\newblock U-mamba: Enhancing long-range dependency for biomedical image segmentation.
\newblock \emph{arXiv preprint arXiv:2401.04722}, 2024.

\bibitem[Nguyen et~al.(2022)Nguyen, Goel, Gu, Downs, Shah, Dao, Baccus, and R{\'e}]{nguyen2022s4nd}
Eric Nguyen, Karan Goel, Albert Gu, Gordon Downs, Preey Shah, Tri Dao, Stephen Baccus, and Christopher R{\'e}.
\newblock S4nd: Modeling images and videos as multidimensional signals with state spaces.
\newblock \emph{NeurIPS}, 2022.

\bibitem[Oquab et~al.(2023)Oquab, Darcet, Moutakanni, Vo, Szafraniec, Khalidov, Fernandez, Haziza, Massa, El-Nouby, Howes, Huang, Xu, Sharma, Li, Galuba, Rabbat, Assran, Ballas, Synnaeve, Misra, Jegou, Mairal, Labatut, Joulin, and Bojanowski]{dinov2}
Maxime Oquab, Timothée Darcet, Theo Moutakanni, Huy~V. Vo, Marc Szafraniec, Vasil Khalidov, Pierre Fernandez, Daniel Haziza, Francisco Massa, Alaaeldin El-Nouby, Russell Howes, Po-Yao Huang, Hu~Xu, Vasu Sharma, Shang-Wen Li, Wojciech Galuba, Mike Rabbat, Mido Assran, Nicolas Ballas, Gabriel Synnaeve, Ishan Misra, Herve Jegou, Julien Mairal, Patrick Labatut, Armand Joulin, and Piotr Bojanowski.
\newblock Dinov2: Learning robust visual features without supervision, 2023.

\bibitem[Pei et~al.(2024)Pei, Huang, and Xu]{pei2024efficientvmamba}
Xiaohuan Pei, Tao Huang, and Chang Xu.
\newblock Efficientvmamba: Atrous selective scan for light weight visual mamba.
\newblock \emph{arXiv preprint arXiv:2403.09977}, 2024.

\bibitem[Radford et~al.(2021)Radford, Kim, Hallacy, Ramesh, Goh, Agarwal, Sastry, Askell, Mishkin, Clark, et~al.]{clip}
Alec Radford, Jong~Wook Kim, Chris Hallacy, Aditya Ramesh, Gabriel Goh, Sandhini Agarwal, Girish Sastry, Amanda Askell, Pamela Mishkin, Jack Clark, et~al.
\newblock Learning transferable visual models from natural language supervision.
\newblock In \emph{ICML}, 2021.

\bibitem[Radosavovic et~al.(2020)Radosavovic, Kosaraju, Girshick, He, and Doll{\'a}r]{radosavovic2020designing}
Ilija Radosavovic, Raj~Prateek Kosaraju, Ross Girshick, Kaiming He, and Piotr Doll{\'a}r.
\newblock Designing network design spaces.
\newblock In \emph{CVPR}, 2020.

\bibitem[Ren et~al.(2015)Ren, He, Girshick, and Sun]{NIPS2015_14bfa6bb}
Shaoqing Ren, Kaiming He, Ross Girshick, and Jian Sun.
\newblock Faster r-cnn: Towards real-time object detection with region proposal networks.
\newblock In \emph{NeurIPS}, 2015.

\bibitem[Ren et~al.(2022)Ren, Zhou, He, Feng, and Wang]{stunned}
Sucheng Ren, Daquan Zhou, Shengfeng He, Jiashi Feng, and Xinchao Wang.
\newblock Shunted self-attention via multi-scale token aggregation.
\newblock In \emph{Proceedings of the IEEE/CVF Conference on Computer Vision and Pattern Recognition}, pages 10853--10862, June 2022.

\bibitem[Ronneberger et~al.(2015)Ronneberger, Fischer, and Brox]{unet}
Olaf Ronneberger, Philipp Fischer, and Thomas Brox.
\newblock {U-Net: Convolutional Networks for Biomedical Image Segmentation}.
\newblock In \emph{MICCAI}, 2015.

\bibitem[Ruan and Xiang(2024)]{Ruan2024Feb_vm_unet}
Jiacheng Ruan and Suncheng Xiang.
\newblock Vm-unet: Vision mamba unet for medical image segmentation.
\newblock \emph{arXiv preprint arXiv:2402.02491}, 2024.

\bibitem[Sandler et~al.(2018)Sandler, Howard, Zhu, Zhmoginov, and Chen]{mobilenetv2}
Mark Sandler, Andrew Howard, Menglong Zhu, Andrey Zhmoginov, and Liang-Chieh Chen.
\newblock {MobileNetV2: Inverted Residuals and Linear Bottlenecks}.
\newblock In \emph{CVPR}, 2018.

\bibitem[Shi(2024)]{Shi2023Nov_transnext}
Dai Shi.
\newblock {TransNeXt: Robust Foveal Visual Perception for Vision Transformers}.
\newblock In \emph{CVPR}, 2024.

\bibitem[Simonyan and Zisserman(2015)]{vgg}
Karen Simonyan and Andrew Zisserman.
\newblock Very deep convolutional networks for large-scale image recognition.
\newblock In \emph{ICLR}, 2015.

\bibitem[Smith et~al.(2023)Smith, Warrington, and Linderman]{Smith2022AugS5}
Jimmy T.~H. Smith, Andrew Warrington, and Scott~W. Linderman.
\newblock {Simplified State Space Layers for Sequence Modeling}.
\newblock In \emph{ICLR}, 2023.

\bibitem[Srinivas et~al.(2021)Srinivas, Lin, Parmar, Shlens, Abbeel, and Vaswani]{BoT_srinivas2021}
Aravind Srinivas, Tsung-Yi Lin, Niki Parmar, Jonathon Shlens, Pieter Abbeel, and Ashish Vaswani.
\newblock Bottleneck transformers for visual recognition.
\newblock \emph{arXiv preprint arXiv:2101.11605}, 2021.

\bibitem[Sun et~al.(2023)Sun, Dong, Huang, Ma, Xia, Xue, Wang, and Wei]{Sun2023Jul}
Yutao Sun, Li~Dong, Shaohan Huang, Shuming Ma, Yuqing Xia, Jilong Xue, Jianyong Wang, and Furu Wei.
\newblock Retentive network: A successor to transformer for large language models.
\newblock \emph{arXiv preprint arXiv:2307.08621}, 2023.

\bibitem[Szegedy et~al.(2015)Szegedy, Liu, Jia, Sermanet, Reed, Anguelov, Erhan, Vanhoucke, and Rabinovich]{inception}
Christian Szegedy, Wei Liu, Yangqing Jia, Pierre Sermanet, Scott Reed, Dragomir Anguelov, Dumitru Erhan, Vincent Vanhoucke, and Andrew Rabinovich.
\newblock {Going deeper with convolutions}.
\newblock In \emph{CVPR}, 2015.

\bibitem[Tan and Le(2019)]{efficientnet_pmlr_tan_19}
Mingxing Tan and Quoc Le.
\newblock {EfficientNet: Rethinking Model Scaling for Convolutional Neural Networks}.
\newblock In Kamalika Chaudhuri and Ruslan Salakhutdinov, editors, \emph{Proceedings of the 36th International Conference on Machine Learning}, pages 6105--6114, Long Beach, California, USA, June 2019. PMLR.

\bibitem[Tolstikhin et~al.(2021)Tolstikhin, Houlsby, Kolesnikov, Beyer, Zhai, Unterthiner, Yung, Steiner, Keysers, Uszkoreit, et~al.]{tolstikhin2021mlp}
Ilya~O Tolstikhin, Neil Houlsby, Alexander Kolesnikov, Lucas Beyer, Xiaohua Zhai, Thomas Unterthiner, Jessica Yung, Andreas Steiner, Daniel Keysers, Jakob Uszkoreit, et~al.
\newblock Mlp-mixer: An all-mlp architecture for vision.
\newblock \emph{Advances in Neural Information Processing Systems}, 34:\penalty0 24261--24272, 2021.

\bibitem[Touvron et~al.(2021)Touvron, Cord, Douze, Massa, Sablayrolles, and J\'egou]{DeiT_touvron2020}
Hugo Touvron, Matthieu Cord, Matthijs Douze, Francisco Massa, Alexandre Sablayrolles, and Herv\'e J\'egou.
\newblock Training data-efficient image transformers \& distillation through attention.
\newblock In \emph{Proceedings of the 38th International Conference on Machine Learning}, 2021.

\bibitem[Touvron et~al.(2023)Touvron, Lavril, Izacard, Martinet, Lachaux, Lacroix, Rozi{\`e}re, Goyal, Hambro, Azhar, et~al.]{touvron2023llama}
Hugo Touvron, Thibaut Lavril, Gautier Izacard, Xavier Martinet, Marie-Anne Lachaux, Timoth{\'e}e Lacroix, Baptiste Rozi{\`e}re, Naman Goyal, Eric Hambro, Faisal Azhar, et~al.
\newblock Llama: Open and efficient foundation language models.
\newblock \emph{arXiv preprint arXiv:2302.13971}, 2023.

\bibitem[Vaswani et~al.(2017)Vaswani, Shazeer, Parmar, Uszkoreit, Jones, Gomez, Kaiser, and Polosukhin]{Transformer_NIPS2017_Vaswani}
Ashish Vaswani, Noam Shazeer, Niki Parmar, Jakob Uszkoreit, Llion Jones, Aidan~N Gomez, ukasz Kaiser, and Illia Polosukhin.
\newblock {Attention is All you Need}.
\newblock In I~Guyon, U~V Luxburg, S~Bengio, H~Wallach, R~Fergus, S~Vishwanathan, and R~Garnett, editors, \emph{Advances in Neural Information Processing Systems}. Curran Associates, Inc., 2017.

\bibitem[Wang et~al.(2017)Wang, Jiang, Qian, Yang, Li, Zhang, Wang, and Tang]{residual_attention}
Fei Wang, Mengqing Jiang, Chen Qian, Shuo Yang, Cheng Li, Honggang Zhang, Xiaogang Wang, and Xiaoou Tang.
\newblock {Residual Attention Network for Image Classification}.
\newblock In \emph{CVPR}, 2017.

\bibitem[Wang et~al.(2023)Wang, Kim, Feris, and Betke]{wang2023cdac}
Kaihong Wang, Donghyun Kim, Rogerio Feris, and Margrit Betke.
\newblock Cdac: Cross-domain attention consistency in transformer for domain adaptive semantic segmentation.
\newblock In \emph{Proceedings of the IEEE/CVF International Conference on Computer Vision}, pages 11519--11529, 2023.

\bibitem[Wang et~al.(2020)Wang, Wu, Zhu, Li, Zuo, and Hu]{ECA_wang2020}
Qilong Wang, Banggu Wu, Pengfei Zhu, Peihua Li, Wangmeng Zuo, and Qinghua Hu.
\newblock Eca-net: Efficient channel attention for deep convolutional neural networks.
\newblock In \emph{The IEEE Conference on Computer Vision and Pattern Recognition}, 2020.

\bibitem[Wang et~al.(2021)Wang, Xie, Li, Fan, Song, Liang, Lu, Luo, and Shao]{PVT_wang2021}
Wenhai Wang, Enze Xie, Xiang Li, Deng-Ping Fan, Kaitao Song, Ding Liang, Tong Lu, Ping Luo, and Ling Shao.
\newblock Pyramid vision transformer: A versatile backbone for dense prediction without convolutions.
\newblock In \emph{Proceedings of the IEEE/CVF International Conference on Computer Vision}, 2021.

\bibitem[Wang et~al.(2022)Wang, Xie, Li, Fan, Song, Liang, Lu, Luo, and Shao]{pvtv2}
Wenhai Wang, Enze Xie, Xiang Li, Deng-Ping Fan, Kaitao Song, Ding Liang, Tong Lu, Ping Luo, and Ling Shao.
\newblock Pvtv2: Improved baselines with pyramid vision transformer.
\newblock \emph{Computational Visual Media}, 2022.

\bibitem[Wang and Ma(2024)]{Wang2024Feb_semi_mamba_unet}
Ziyang Wang and Chao Ma.
\newblock Semi-mamba-unet: Pixel-level contrastive cross-supervised visual mamba-based unet for semi-supervised medical image segmentation.
\newblock \emph{arXiv preprint arXiv:2402.07245}, 2024.

\bibitem[Wang et~al.(2024)Wang, Zheng, Zhang, Cui, and Li]{Wang2024Feb_mamba_unet}
Ziyang Wang, Jian-Qing Zheng, Yichi Zhang, Ge~Cui, and Lei Li.
\newblock Mamba-unet: Unet-like pure visual mamba for medical image segmentation.
\newblock \emph{arXiv preprint arXiv:2402.05079}, 2024.

\bibitem[Williams et~al.(2007)Williams, Lawrence, et~al.]{lti}
Robert~L Williams, Douglas~A Lawrence, et~al.
\newblock \emph{Linear state-space control systems}.
\newblock John Wiley \& Sons, 2007.

\bibitem[Woo et~al.(2023)Woo, Debnath, Hu, Chen, Liu, Kweon, and Xie]{convnextv2}
Sanghyun Woo, Shoubhik Debnath, Ronghang Hu, Xinlei Chen, Zhuang Liu, In~So Kweon, and Saining Xie.
\newblock {ConvNeXt V2: Co-designing and Scaling ConvNets with Masked Autoencoders}.
\newblock In \emph{CVPR}, 2023.

\bibitem[Wu et~al.(2021)Wu, Xiao, Codella, Liu, Dai, Yuan, and Zhang]{CvT_Wu_2021tw}
Haiping Wu, Bin Xiao, Noel Codella, Mengchen Liu, Xiyang Dai, Lu~Yuan, and Lei Zhang.
\newblock Cvt: Introducing convolutions to vision transformers.
\newblock In \emph{Proceedings of the IEEE/CVF International Conference on Computer Vision}, 2021.

\bibitem[Wu et~al.(2019)Wu, Shen, and Van Den~Hengel]{wu2019wider}
Zifeng Wu, Chunhua Shen, and Anton Van Den~Hengel.
\newblock Wider or deeper: Revisiting the resnet model for visual recognition.
\newblock \emph{Pattern Recognition}, 90:\penalty0 119--133, 2019.

\bibitem[Xiao et~al.(2018)Xiao, Liu, Zhou, Jiang, and Sun]{xiao2018unified}
Tete Xiao, Yingcheng Liu, Bolei Zhou, Yuning Jiang, and Jian Sun.
\newblock Unified perceptual parsing for scene understanding.
\newblock In \emph{Proceedings of the European conference on computer vision}, pages 418--434, 2018.

\bibitem[Xie et~al.(2016)Xie, Girshick, Doll{\ifmmode\acute{a}\else\'{a}\fi}r, Tu, and He]{resnext}
Saining Xie, Ross Girshick, Piotr Doll{\ifmmode\acute{a}\else\'{a}\fi}r, Zhuowen Tu, and Kaiming He.
\newblock {Aggregated Residual Transformations for Deep Neural Networks}.
\newblock In \emph{CVPR}, 2016.

\bibitem[Xie et~al.(2017)Xie, Girshick, Doll{\'a}r, Tu, and He]{xie2017aggregated}
Saining Xie, Ross Girshick, Piotr Doll{\'a}r, Zhuowen Tu, and Kaiming He.
\newblock Aggregated residual transformations for deep neural networks.
\newblock In \emph{Proceedings of the IEEE conference on computer vision and pattern recognition}, pages 1492--1500, 2017.

\bibitem[Xu et~al.(2021{\natexlab{a}})Xu, Xu, Chang, and Tu]{xu2021co}
Weijian Xu, Yifan Xu, Tyler Chang, and Zhuowen Tu.
\newblock Co-scale conv-attentional image transformers.
\newblock In \emph{Proceedings of the IEEE/CVF International Conference on Computer Vision}, pages 9981--9990, 2021{\natexlab{a}}.

\bibitem[Xu et~al.(2021{\natexlab{b}})Xu, Zhang, Zhang, and Tao]{xu2021vitae}
Yufei Xu, Qiming Zhang, Jing Zhang, and Dacheng Tao.
\newblock Vitae: Vision transformer advanced by exploring intrinsic inductive bias.
\newblock \emph{Advances in Neural Information Processing Systems}, 34:\penalty0 28522--28535, 2021{\natexlab{b}}.

\bibitem[Xue et~al.(2022)Xue, Shi, Wei, Lou, Liu, and You]{xue2022go}
Fuzhao Xue, Ziji Shi, Futao Wei, Yuxuan Lou, Yong Liu, and Yang You.
\newblock Go wider instead of deeper.
\newblock In \emph{Proceedings of the AAAI Conference on Artificial Intelligence}, volume~36, pages 8779--8787, 2022.

\bibitem[Yang et~al.(2023)Yang, Xu, Mello, Crowley, and Wang]{yang2023gpvit}
Chenhongyi Yang, Jiarui Xu, Shalini~De Mello, Elliot~J. Crowley, and Xiaolong Wang.
\newblock {GPViT: A High Resolution Non-Hierarchical Vision Transformer with Group Propagation}.
\newblock In \emph{ICLR}, 2023.

\bibitem[Yang et~al.(2021)Yang, Li, Zhang, Dai, Xiao, Yuan, and Gao]{focal}
Jianwei Yang, Chunyuan Li, Pengchuan Zhang, Xiyang Dai, Bin Xiao, Lu~Yuan, and Jianfeng Gao.
\newblock Focal self-attention for local-global interactions in vision transformers.
\newblock In \emph{NeurIPS}, 2021.

\bibitem[Yun et~al.(2019)Yun, Han, Oh, Chun, Choe, and Yoo]{yun2019cutmix}
Sangdoo Yun, Dongyoon Han, Seong~Joon Oh, Sanghyuk Chun, Junsuk Choe, and Youngjoon Yoo.
\newblock Cutmix: Regularization strategy to train strong classifiers with localizable features.
\newblock In \emph{Proceedings of the IEEE/CVF international conference on computer vision}, pages 6023--6032, 2019.

\bibitem[Zhai et~al.(2022)Zhai, Kolesnikov, Houlsby, and Beyer]{Zhai2021Jun_scaling_vits}
Xiaohua Zhai, Alexander Kolesnikov, Neil Houlsby, and Lucas Beyer.
\newblock {Scaling Vision Transformers}.
\newblock In \emph{CVPR}, 2022.

\bibitem[Zhang et~al.(2017)Zhang, Cisse, Dauphin, and Lopez-Paz]{zhang2017mixup}
Hongyi Zhang, Moustapha Cisse, Yann~N Dauphin, and David Lopez-Paz.
\newblock mixup: Beyond empirical risk minimization.
\newblock \emph{arXiv preprint arXiv:1710.09412}, 2017.

\bibitem[Zhang et~al.(2021)Zhang, Dai, Yang, Xiao, Yuan, Zhang, and Gao]{ViL_Zhang_2021tu}
Pengchuan Zhang, Xiyang Dai, Jianwei Yang, Bin Xiao, Lu~Yuan, Lei Zhang, and Jianfeng Gao.
\newblock {Multi-Scale Vision Longformer: A New Vision Transformer for High-Resolution Image Encoding}.
\newblock \emph{arXiv.org}, March 2021.

\bibitem[Zhong et~al.(2020)Zhong, Zheng, Kang, Li, and Yang]{zhong2020random}
Zhun Zhong, Liang Zheng, Guoliang Kang, Shaozi Li, and Yi~Yang.
\newblock Random erasing data augmentation.
\newblock In \emph{Proceedings of the AAAI conference on artificial intelligence}, 2020.

\bibitem[Zhou et~al.(2017)Zhou, Zhao, Puig, Fidler, Barriuso, and Torralba]{ADK20K}
Bolei Zhou, Hang Zhao, Xavier Puig, Sanja Fidler, Adela Barriuso, and Antonio Torralba.
\newblock Scene parsing through ade20k dataset.
\newblock In \emph{Proceedings of the IEEE Conference on Computer Vision and Pattern Recognition}, July 2017.

\bibitem[Zhu et~al.(2024)Zhu, Liao, Zhang, Wang, Liu, and Wang]{zhu2024vision}
Lianghui Zhu, Bencheng Liao, Qian Zhang, Xinlong Wang, Wenyu Liu, and Xinggang Wang.
\newblock Vision mamba: Efficient visual representation learning with bidirectional state space model.
\newblock \emph{arXiv preprint arXiv:2401.09417}, 2024.

\end{thebibliography}
\clearpage

\appendix

\section{More Experiments}

\subsection{COCO Object Detection using RetinaNet}

We report COCO RetinaNet object detection in Table~\ref{table:retinanetSmall}. Similar to Mask R-CNN, whose results are reported in the main paper, \ourmethod~also performs well with the single-stage RetinaNet object detector. For example, with only half the model size and similar FLOPs, \ourmethod-L1 achieves 0.2 higher AP than Swin-Tiny.

\begin{table}[h]
\centering
\centering
\caption{\small RetinaNet object detection on MS COCO \textit{mini-val} with 1$\times$ schedule. FLOPs are computed using input size  1280$\times$800.}
\label{table:retinanetSmall}
\vspace{3mm}
\begin{adjustbox}{max width=0.7\textwidth}
\begin{tabular}{lccc|ccc}
\hline
Backbone & Hierarchical &  Params &   FLOPs & $AP^{bb}$  & $AP^{bb}_{50}$ &  $AP^{bb}_{75}$ \\
\hline
\multicolumn{6}{l}{CNN} \\
\hline
ResNeXt101-32x4d~\cite{xie2017aggregated} & \checkmark & 56M  & 319G & 39.9 & - & - \\
ResNeXt101-64×4d~\cite{xie2017aggregated} & \checkmark & 95M  & 413G & 41.0 & - & - \\
\hline
\multicolumn{5}{l}{Transformer} \\
\hline
Swin-Tiny~\cite{Swin_Liu_2021tq} & \checkmark & 38M & 245G & 41.5 & - & - \\
Swin-Small~\cite{Swin_Liu_2021tq} & \checkmark & 60M & 335G & 44.5 & - & - \\

Focal-Tiny~\cite{focal} & \checkmark & 39M & 265G & 43.7 & - & -  \\
Focal-Small~\cite{focal} & \checkmark & 62M & 367G & 45.6 & - & -  \\

PVT-Small~\cite{PVT_wang2021} & \checkmark & 34M & - & 40.4 & 61.3 & 43.0 \\
PVT-Medium~\cite{PVT_wang2021} & \checkmark & 54M & - & 41.9 & 63.1 & 44.3 \\
PVT-Large~\cite{PVT_wang2021} & \checkmark & 71M & - & 42.6 & 63.7 & 45.4 \\

\hline
\multicolumn{5}{l}{State Space Modeling} \\
\hline
EfficientVMamba-T~\cite{pei2024efficientvmamba} &\checkmark & 13M & - & 37.5 & 57.8 & 39.6 \\
EfficientVMamba-S~\cite{pei2024efficientvmamba} &\checkmark & 19M & - & 39.1 & 60.3 & 41.2 \\
EfficientVMamba-B~\cite{pei2024efficientvmamba} &\checkmark & 44M & - & 42.8 & 63.9 & 45.8 \\
\hline
\ourmethod-Adapter-L1 & \xmark & 19M & 250G & 41.7 & 62.1 & 44.4 \\
\ourmethod-Adapter-L2& \xmark  & 40M & 392G & 43.9 & 64.9 & 47.0 \\
\ourmethod-Adapter-L3& \xmark  & 67M & 478G & 44.8 & 66.0 & 47.9 \\
\hline 
\end{tabular}
\end{adjustbox}
\end{table}

\subsection{Ablation Studies and Discussions}

\textbf{Setting:} Here, we conduct ablation studies to test our model designs and to gain a deeper understanding of the proposed method. We use our L1 model, with less than 10M parameters, for most experiments. The models are all pre-trained on ImageNet-1K following the same training settings described in the main paper. \\

\begin{table}[h]
\centering
\caption{\small Ablation study of model depth v.s. width on ImageNet-1K.. }
\label{table:depthwidth}
\vspace{2mm}
\begin{adjustbox}{max width=.7\textwidth}
\begin{tabular}{@{\hskip 10pt}c@{\hskip 10pt}c@{\hskip 10pt}|@{\hskip 10pt}r@{\hskip 10pt}r@{\hskip 10pt}c@{\hskip 10pt}}
\hline
Depth & Width   &  Params & FLOPs  & Top-1  \\
\hline
6 &    376  & 7.3 M &  2.5 G  &  74.6 \\
12 &    272  & 7.5 M&  2.7 G  &  76.8  \\
24 &    192  & 7.3 M &  3.0 G  &  77.9 \\
36 &    156   &7.2 M &  3.3 G  &  77.9 \\
\hline
\end{tabular}
\end{adjustbox}

\end{table}

\noindent\textbf{Depth v.s. Width}
When designing neural architectures for a given parameter count, it's usually important to find a good balance between the network's depth, i.e., the number of layers, and its width, i.e., the feature dimensions. While this problem was studied for existing architectures~\cite{wu2019wider,xue2022go}, it is still unclear whether the previous conclusions are applicable to vision SSMs. In Table~\ref{table:depthwidth}, we study the depth and width trade-off of the proposed \ourmethod. Firstly, the results show that deeper models tend to perform better than shallow ones. For example, when the parameter count is around 7.4M, the 12-layer model achieves 2.2\% higher ImageNet top-1 accuracy than the 6-layer counterparts, and the 24-layer model is further 1.1\% higher than the 12-layer one. However, when we further increase the depth to 36 while reducing the width accordingly, the top-1 accuracy remains similar. On the other hand, we also notice that deeper models are less efficient than shallower but wider models. For instance, the 24-layer model is 0.3G FLOPs higher than the 12-layer model. These results suggest the necessity of a good balance between network depth and width.

\begin{figure*}[t]
    \centering
    \includegraphics[width=0.8\textwidth]{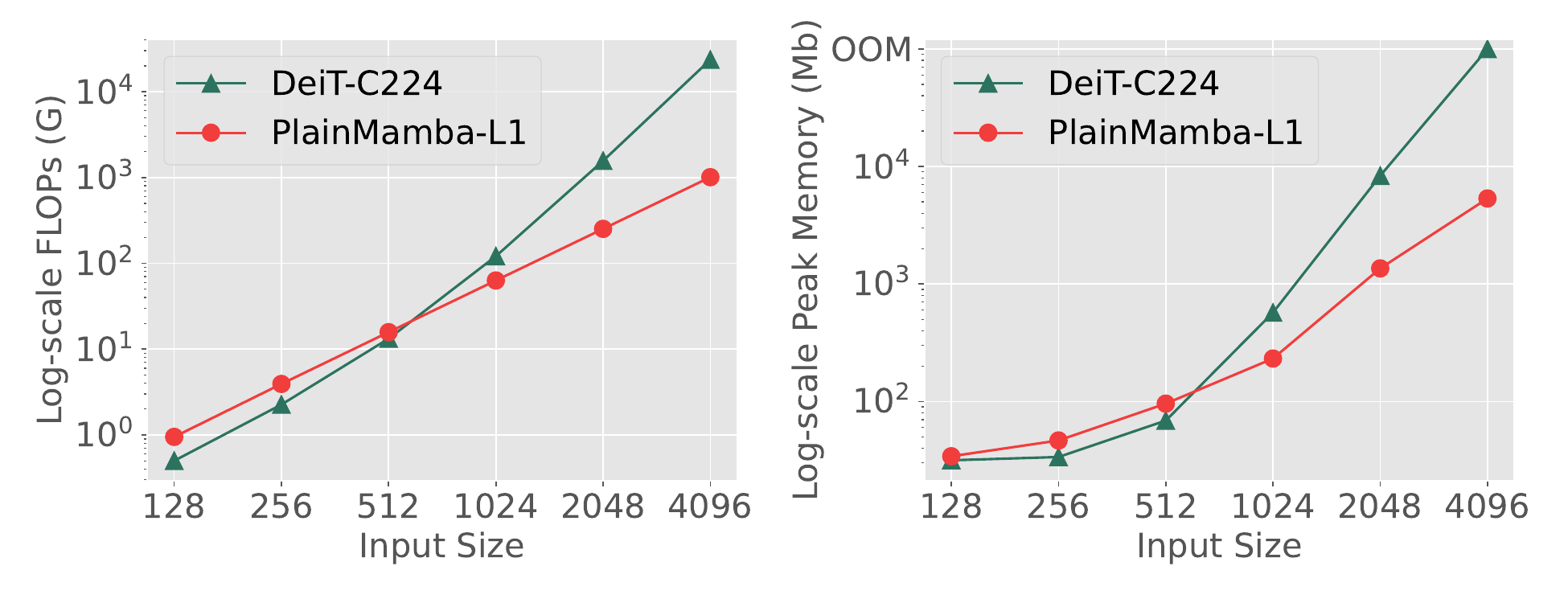}
    \captionsetup{font={small}}
\caption{\small Efficiency comparison between \ourmethod~and DeiT. We modify the DeiT-Tiny model by changing its channel number to 224, resulting in a similar-size model (7.4M) to \ourmethod-L1. The peak memory is measured using a batch size of 1. }
\label{fig:flops}

\vspace{-2mm}
\end{figure*}

\begin{wraptable}{R}{0.5\textwidth}
\centering
\caption{\small Ablation study of model depth v.s. width on ImageNet-1K.}
\label{table:block}
\vspace{3mm}
\begin{adjustbox}{max width=0.5\textwidth}
\begin{tabular}{l@{\hskip 13pt}@{\hskip 13pt}r@{\hskip 13pt}r@{\hskip 13pt}r@{\hskip 13pt}}
\hline
Method    &  Params & FLOPs  & Top-1  \\
\hline
VisionMamba~\cite{zhu2024vision} &    7.8M  & 1.3G &  74.4 \\
VMamba~\cite{liu2024vmamba} &    7.3M  & 3.0G &  77.1 \\
\hline
Ours & 7.3M  & 3.0G &  77.9  \\
\hline
\end{tabular}
\end{adjustbox}
\end{wraptable}

\noindent\textbf{\ourmethod~Block Design.} 
Here, we test different designs of the \ourmethod~block by comparing it with the block designs in Vision Mamba~\cite{zhu2024vision} and VMamba~\cite{liu2024vmamba}. For a fair comparison, we use the same model depth and width settings for all designs and train all models with the same training recipe. We also remove the \texttt{CLS} tokens from the Vision Mamba~\cite{zhu2024vision} block and use the global averaging pooling as an alternative. We report the results in Table~\ref{table:block}. We can see that our design achieves the best results. Specifically, Vision Mamba only achieves a 74.4\% ImageNet accuracy, which is 3.5\% lower than ours. We also notice that the model with a Vision Mamba block is inferior to the original Vision Mamba model, which is caused by the removal of the \texttt{CLS} token. These results suggest that our model still retains its ability when the \texttt{CLS} token is absent. Also, our design performs better than the VMamba block~\cite{liu2024vmamba} with a 0.8\% accuracy advantage, indicating that the improvements come from our proposed Continuous 2D Scanning and Direction-aware Updating, which validate the effectiveness of our proposed techniques in adapting SSM for 2D images. \\

\begin{wraptable}{R}{0.5\textwidth}
\centering
\caption{\small Comparison of decomposed FLOPs between DeiT and \ourmethod.}
\label{table:flopsDecompose}
\vspace{2mm}
\begin{adjustbox}{max width=0.5\textwidth}
\begin{tabular}{c|c|@{\hskip 5pt}r@{\hskip 5pt}r@{\hskip 5pt}}
\hline
Resolution & Part  &  DeiT-C224 & \ourmethod-L1  \\
\hline
\multirow{3}{*}{128$\times$128} & Token Mixing & 0.18G & 0.34G \\
                     & Channel Mixing & 0.31G & 0.33G \\
                     & Others & 0.01G & 0.30G\\
\hline
\multirow{3}{*}{4096$\times$4096} & Token Mixing & 23244G & 350G \\
                     & Channel Mixing & 315G & 348G \\
                     & Others & 12G & 311G \\
\hline
\end{tabular}
\end{adjustbox}
\end{wraptable}

\noindent\textbf{Efficiency Comparison with ViT}
One particular advantage of SSMs, e.g., Mamba, is their ability to capture global information while maintaining efficiency. in Figure~\ref{fig:flops}, we compare the \ourmethod's efficiency with the vision transformer. Specifically, to ensure a fair comparison, we create a DeiT model with channel numbers of 224, resulting in a model with 7.4M parameters, which is used to compare with \ourmethod-L1. Specifically, we compare the model FLOPs and the peak inference memory using inputs of different sizes. The results show that our model is able to keep the computation cost low when the input size is scaled up to high resolutions, e.g., 4096$\times$4096. However, DeiT's FLOPs and memory consumption increase rapidly when using such high-resolution inputs. On the other hand, we also notice that our model's efficiency is inferior to the similar-sized DeiT when using low-resolution images, e.g., 128$\times$128. To further investigate such a difference in their efficiency, we decompose their FLOPs into three parts~\cite{tolstikhin2021mlp}: 1) \textit{token mixing}, 2) \textit{channel mixing}, and 3) others. Specifically, token mixing refers to the multi-head attention part in DeiT and the selective scanning part in \ourmethod, and channel mixing refers to the feed-forward network in DeiT and the input \& output projection in \ourmethod. We report the results in Table~\ref{table:flopsDecompose}. These suggest that \ourmethod's FLOPs are evenly distributed across the three parts in low and high resolutions. On the contrary, when using 128$\times$128 inputs, DeiT's FLOPs are dominated by channel-mixing and the other two parts are negligible. However, because of the quadratic complexity of self-attention operation, DeiT's FLOPs in token mixing grow to 23T when using 4096$\times$4096 inputs, 23 times more expensive than \ourmethod. These results verify \ourmethod's high efficiency for high-resolution inputs.

\end{document}